\documentclass[article,authoryear,jmlmc]{beg_32}             

\usepackage[hang]{footmisc}
\usepackage{hyperref}
\fancypagestyle{plain}{%
  \fancyhf{}
  \fancyhead[R]{\small {\it \jname}, x(x):\thepage--\pageref{LastPage} (\myyear\today)}
  \fancyfoot[R]{\small\bf\thepage }
  \fancyfoot[L]{\fottitle}
  }
\renewcommand{\dmyy}{25}
\renewcommand{\myyear}{2025}
\renewcommand{\today}{}
\begin{document}

\volume{Volume x, Issue x, \myyear\today}
\title{Federated Learning on Stochastic Neural Networks}
\titlehead{FedStNN}
\authorhead{Jingqiao Tang, Ryan Bausback, Feng Bao, \& Richard Archibald}
\corrauthor[1]{Jingqiao Tang}
\author[1]{Ryan Bausback}
\author[1]{Feng Bao}
\author[2]{Richard. Archibald}
\corremail{jt21v@fsu.edu}
\corraddress{Department of Mathematics at Florida State University, Tallahassee, Florida, 32304}
\address[1]{Department of Mathematics at Florida State University, Tallahassee, Florida, 32304}
\address[2]{Division of Computer Science and Mathematics, Oak Ridge National Laboratory, Oak Ridge, Tennessee, 37830}

\dataO{05/05/2025}
\dataF{mm/dd/yyyy}

\abstract{Federated learning is a machine learning paradigm that leverages edge computing on client devices to optimize models while maintaining user privacy by ensuring that local data remains on the device. However, since all data is collected by clients, federated learning is susceptible to latent noise in local datasets. Factors such as limited measurement capabilities or human errors may introduce inaccuracies in client data. To address this challenge, we propose the use of a stochastic neural network as the local model within the federated learning framework. Stochastic neural networks not only facilitate the estimation of the true underlying states of the data but also enable the quantification of latent noise. We refer to our federated learning approach, which incorporates stochastic neural networks as local models, as Federated stochastic neural networks. We will present numerical experiments demonstrating the performance and effectiveness of our method, particularly in handling non-independent and identically distributed data.
}


\maketitle

\section{Introduction} \label{sec:intro}
The fundamental principles of federated learning can be traced back to earlier advancements in distributed computing and privacy-preserving machine learning techniques. Before federated learning was introduced in \cite{DBLP:journals/corr/McMahanMRA16}, distributed machine learning primarily focused on executing training processes in parallel across multiple nodes within a data center. Notable frameworks, such as MapReduce (\cite{mapreduce}) and AllReduce, were designed to aggregate data from different computational units, perform global aggregation using predefined operators, and subsequently redistribute the outcomes to all participating units. However, these methods operated under the assumption that data could be freely exchanged among computational nodes — a premise that became increasingly infeasible because of the emergence of stringent privacy regulations and growing concerns over the security of user data. 

Privacy-preserving machine learning (PPML) is a subfield of machine learning that focuses on developing methodologies to train models while safeguarding sensitive information from unauthorized access. The primary objective of PPML is to design techniques that protect confidential data for both individuals and organizations. This process generally involves three key steps: (1) identifying potential risks and understanding relevant regulatory requirements, (2) measuring vulnerabilities and the success of attacks, and (3) implementing strategies to mitigate these risks. Among the most widely adopted approaches in PPML is Differential Privacy (DP) (\cite{differential_privacy}), which introduces controlled noise to prevent individual data points from being inferred. Another famous technique is Homomorphic Encryption (HE) (\cite{homomorphic_encryption}), which enables computations to be performed directly on encrypted data without requiring decryption, thereby ensuring data confidentiality throughout the learning process.

Traditional machine learning algorithms generally require developers to collect data from users before applying an optimization/learning process. This data collection step usually raises significant concerns regarding individual privacy. Moreover, due to the varying scale of personal datasets, transmitting large volumes of personal data may cost a huge amount of communication resources. To address this concern, \textbf{Federated Learning}, also known as \textbf{Centralized Federated Learning} (\textbf{CFL}), was introduced in \cite{DBLP:journals/corr/McMahanMRA16}. The idea is to set up a central server that can communicate with all users/clients. Instead of sharing raw data directly with the server, each client will train the model individually and upload the model parameters to the server. 

In a general federated setup, it will assume to have $K$ clients, each corresponded with a fixed local training dateset $P_k$, with $n_k = |P_k|$ representing the cardinality of set $P_k$(number of data). At the beginning of each round, a random selection process will choose a fraction $CK$ of clients, with constant $C \in [0, 1]$. It has been shown in \cite{DBLP:journals/corr/McMahanMRA16} that setting $C = 0.1$ will in general be the most efficient choice for convergence. 

The server then transmits the current global model state $u_t$ to the selected clients, sharing only the model parameters. Each client then performs local updates based on the received global model and their respective local datasets. These locally updated model parameters are returned to the server, which aggregates the updates to refine the global model to $u_{t + 1}$. It is evident that the CFL approach not only preserves user privacy and enhances computational efficiency, but also significantly reduces communication overhead — particularly when the size of the model parameters is substantially smaller than that of the raw datasets.

\begin{figure}[!h]
    \centering
    \includegraphics[width=0.5\linewidth]{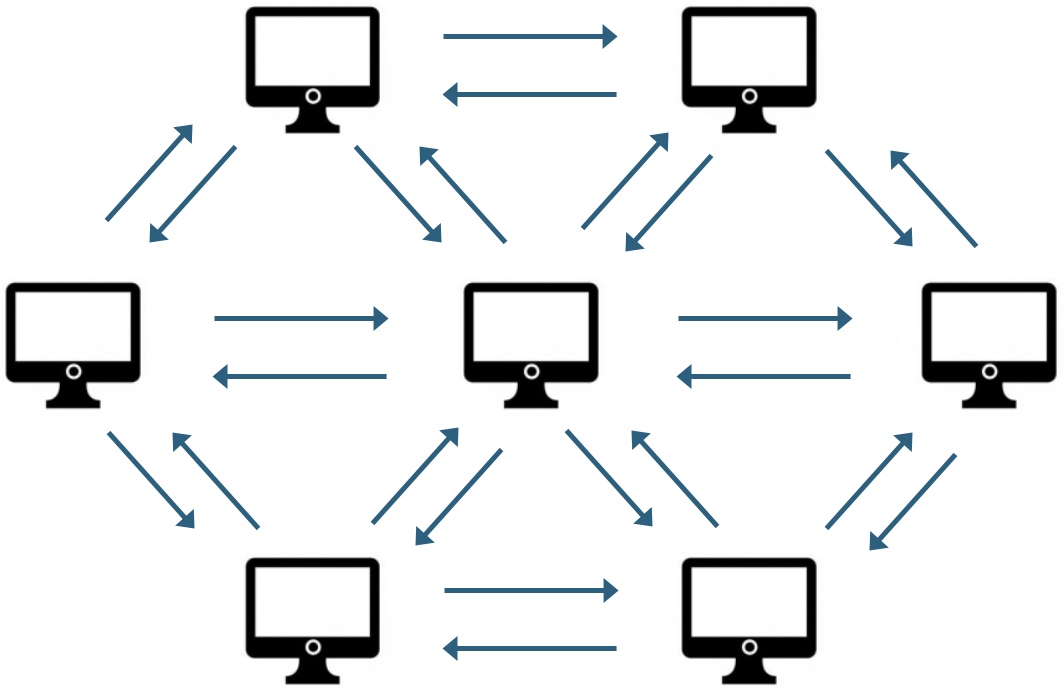}
    \caption{Decentralized Federated Learning}
    \label{fig:DFL}
\end{figure}

In contrast to CFL, which has a central server that collects and aggregates all local models, \textbf{Decentralized Federated Learning}(\textbf{DFL}) methods do not require a central server, and clients can share parameters directly to other clients, as shown in figure \ref{fig:DFL}. Unlike CFL, different DFL methods are introduced based on considering different client network topology. Peer-to-peer FL (\cite{lalitha2019peertopeerfederatedlearninggraphs}), server free FL (\cite{he2020centralserverfreefederated}), serverless FL (\cite{he2021spreadgnnserverlessmultitaskfederated}) are introduced as DFL methods. 

Federated learning methods can also be categorized as \textbf{Horizontal Federated Learning}(\textbf{HFL}), \textbf{Vertical Federated Learning}(\textbf{VFL}), and \textbf{Federated Transfer Learning}(\textbf{FTL}). 

For datasets that have the same feature space but different sample spaces, horizontal federated learning will be present. For example, two regional banks may only have a few customers in common, but given the high similarity of the services, both banks have closed or identical feature spaces. A collaborative deep learning method named Distributed Selected SGD (DSSGD) is introduced in \cite{DSSGD} for such scenarios. In each iteration of the DSSGD algorithm, users download a subset of the global model parameters from the server and replace the corresponding parameters in their local models. Local training is then performed using the individual datasets. Finally, a fraction of local parameters will be selected randomly and uploaded back to the server for global model parameter updates. The DSSGD method shares a similar idea with horizontal federated learning. It has been shown that the federated learning approach with a safe aggregation technique that protects the aggregated user updates can establish a protection on users’ privacy in \cite{Bonawitz}. 

VFL methods are focusing on datasets with similar sample spaces but different feature spaces. An e-commerce business and a bank are examples of companies in the same city with differing business models, yet their customer demographics overlap significantly, as they likely serve a large part of the local population. Both collect customer information such as income, spending habits, and credit scores, though they emphasize different types of data. If both organizations intend to build a model for predicting product purchases using product and user data, their separate datasets would be integrated. The model would then be trained by computing the loss and gradients to combine the information from both parties. In \cite{hardy2017privatefederatedlearningvertically}, a VFL method that can perform logistic regression in the cross-feature space between two datasets by introducing a third-party coordinator is presented. 

Machine learning algorithms generally assume that training and test data share the same distribution and feature space, but this is often not the case in real-world applications. In practice, we often need to study a specific domain of interest while relying on training data from a different domain. Transfer learning offers a solution to this problem by allowing models trained on one dataset to be adapted and applied to another, facilitating knowledge transfer between domains. Federated transfer learning (FTL) is a distinct approach within federated learning, separate from horizontal and vertical federated learning. In FTL, the datasets involved have different feature spaces. For instance, if two parties, A and B, share only a small portion of their feature and sample spaces, a model trained on B can be adapted for A by leveraging this shared data. FTL begins with a model trained on the source domain’s samples and features, then adjusts it for the target domain. This allows the model to handle non-overlapping samples by using knowledge gained from the unique features of the source domain. 

Federated learning (FL) is being widely applied across various fields, enabling privacy-preserving and decentralized AI training. In the healthcare sector, FL facilitates advancements in medical prognosis, diagnosis, and clinical workflow while ensuring data confidentiality (\cite{FL_Health_application}). Financial institutions leverage FL for risk control, marketing, and anti-money laundering, enhancing security without compromising sensitive financial data (\cite{FL_Finance_application}). In the realm of smart devices and the Internet of Things (IoT), FL enhances predictive text functionalities, supports autonomous vehicle development, and optimizes smart home automation (\cite{FL_IoT_application}). Transportation systems benefit from FL through enhanced autonomous driving models (\cite{FL_Autodriving_application}), including vehicle trajectory prediction (\cite{FL_Trajectory_application}), license recognition (\cite{FL_License_application}), motion control (\cite{FL_Motion_applicaiton}), and traffic management (\cite{FL_Traffic_application}). Furthermore, FL contributes to advancements in natural language processing (NLP) and virtual assistants by refining language models while safeguarding user data (\cite{FL_NLP_application}). In the field of education, it  can analyze student performance, personalize learning experiences, and detect learning patterns without compromising sensitive student information (\cite{FL_Education_application}). Finally, FL facilitates data-driven decision-making in smart cities (\cite{FL_SmartCity_application}), smart grids (\cite{FL_SmartGrid_application}), and smart health. This is particularly important for real-time patient monitoring and disease prediction, while maintaining patient data privacy (\cite{FL_realtimehealth_application}). FL therefore ensures scalable and secure artificial intelligence applications.

Federated learning methods effectively leverage edge computing on clients' devices for optimization while preserving user privacy by ensuring that local data remains unshared. However, research indicates that the performance of federated learning can deteriorate significantly when client data is not sampled in an independent and identically distributed (i.i.d.) manner from the overall population (\cite{li2020convergencefedavgnoniiddata}). In other words, the majority of clients' local datasets do not accurately represent the entire population. Beyond the issue of uneven data distribution, federated learning is also susceptible to observation errors. Since all data are exclusively collected by individual clients and never shared, measurement noise may arise due to factors such as instrument limitations, human error, and other inconsistencies. These challenges hinder individuals from accurately assessing the true state of the model. 

To account for uncertainties in federated learning, we aim to determine the probability distribution of the outcome. Specifically, based on prior knowledge—referred to as the prior—and the degree to which the data align with the current parameters—referred to as the likelihood—our objective is to derive the posterior distribution, which represents the probability of the updated parameters given the observed data. This is the approach adopted in the field of Bayesian Statistics, and the method to calculate the posterior is known as the Bayesian Inference. Bayesian Neural Networks (BNNs) are stochastic neural networks trained using a Bayesian approach. It is a widely used method and has the capability to measure the uncertainties (\cite{BNN_intro}). 

In this study, we adopt an alternative neural network model, the Stochastic Neural Network (SNN), originally introduced in \cite{archibald2021backwardsdemethoduncertainty}. By incorporating Gaussian noise into the data, we reformulated local client training as a stochastic optimal control problem (\cite{Bao_EAJAM20, Bao_Control_20}). That is, instead of solving an ordinary differential equation (ODE) using a traditional neural network, each client employs a local SNN to address a stochastic ordinary differential equation (SDE). The details of this method will be elaborated in Section \ref{sec:SNN}. The convergence of the SNN method is also demonstrated in \cite{SNN_Convergence}. 

While both BNNs and SNNs are capable of capturing latent uncertainty in data, SNNs are more suitable for federated learning. One principal rationale for choosing SNN is that BNNs treat parameters or activation functions as random variables, which pose challenges for aggregation in a federated learning framework (see details in Section \ref{sec:FL}). In contrast, SNNs utilize deterministic coefficients. Since SNNs are designed to solve SDEs, they consist of two internal networks within a single model: one dedicated to capturing the drift term in the SDE and the other, the diffusion network, responsible for quantifying data uncertainty (see details in Section \ref{sec:SNN}). Given that all parameters in an SNN are deterministic, this model is more compatible with federated learning.

Accordingly, we propose the \textbf{Federated SNN} (\textbf{FedStNN}), a federated learning algorithm that employs SNNs as local models. We will demonstrate its effectiveness in handling noisy data through numerical experiments. Furthermore, we will evaluate the performance of FedStNN in scenarios involving non-independent and identically distributed (non-iid) data. The remaining of this paper is structured as follows: Section \ref{sec:FL} provides a brief overview of federated learning and BNNs, Section \ref{sec:SNN} details the mathematical formulation and numerical implementation of SNNs, Section \ref{sec:FedSNN} introduces the FedStNN methodology, and Section \ref{sec:experiments} presents experimental results using FedStNN.

\section{Federated Learning} \label{sec:FL}
In this section, we will present a fundamental centralized horizontal federated learning method introduced in \cite{DBLP:journals/corr/McMahanMRA16}. 

\subsection{FedSGD and FedAvg}
\begin{figure}
    \centering
    \includegraphics[width=0.5\linewidth]{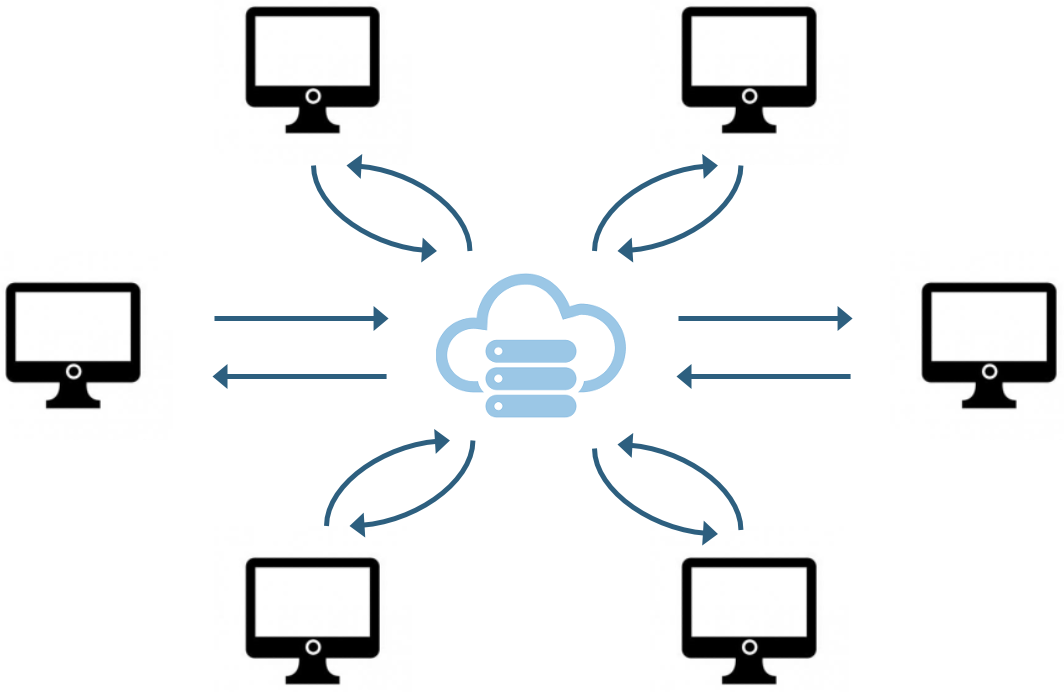}
    \caption{Centralized Federated Learning}
    \label{fig:CFL}
\end{figure}

In a federated learning question, we will assume there to be $K$ clients and every client owns a local model. Each client is able to collect data by himself and can train the local model he owns by using the data he collects. There is a central server (usually a company or institution) which is capable of communicating with all clients. Since we only present a centralized federated model, we assume that clients are forbidden to communicate with each other, as shown in Fig. \ref{fig:CFL}. A global model designated to capture the general trend of the entire population is generated by the central server. 

In data communication, the substantial volume of personal data generated by clients, coupled with the potential instability of network connections (typically the internet), renders the direct transmission of all collected data to the server impractical. Furthermore, for privacy-sensitive information such as health or demographic data, clients often prefer to retain the data locally rather than share it with the server to safeguard their privacy. Therefore, only model parameters, which are typically much smaller in size than the dataset, are transmittable between clients and the server. Additionally, we assume all clients' local models have the same structure as the global model in the server. Specifically, since we are focusing on neural networks, the global model at the server and all local models are neural networks with the same number of layers, the same number of neurons in each layer, and the same activation functions. It is important to note that the authors of federated learning did not explicitly specify the model type as a neural network. The federated learning method presented can be applied to various other models as well. 

For non-convex neural network problems, the goal is to find a model parameter $u \in \mathbb{R}^d$ such that it minimizes the loss function $f(u)$, which is defined as \begin{equation} \label{eqn:loss_function}
    f(u) = \frac{1}{n} \sum_{i = 1}^n f_i(u). 
\end{equation} In traditional machine learning problems, it is usually the case that $f_i(u) = l(x_i, y_i; u)$, where $l$ is the loss function and $l(x, y; u)$ is the loss of prediction on input data $x_i$ and observation $y_i$ with model parameter $u$, and $n$ being the total number of observations. 

Then, in the federated learning setting with $K$ clients and their corresponding local dataset $P_k$, equation \ref{eqn:loss_function} can be rewritten as \begin{equation} \label{eqn:federated_loss_function}
    f(u) = \sum_{k = 1}^K \frac{n_k}{n} F_k(u), 
\end{equation} where $F_k(u)$ is the loss of client $k$ with model parameter $u$, defined by \begin{equation} \label{eqn:federated_loss_function_client}
    F_k(u) = \frac{1}{n_k} \sum_{i \in P_k} f_i(u). 
\end{equation}

If all local data are independent and identically distributed(IID), then we would have $\displaystyle E_{P_k} [F_k(u)] = f(u)$. It is expected that the equation does not hold for non-IID data. 

Given a fixed learning rate $\eta$, each client $k$ can compute its local average gradient $g_k = \nabla F_k(u_t)$ with model parameter $u_t$ in the \textbf{local training step}. Note that \begin{equation} \label{eqn:federated_gradient}
    \nabla f(u_t) = \sum_{k = 1}^K \nabla F_k(u_t) = \sum_{k = 1}^K g_k. 
\end{equation} Then, in the \textbf{federated aggregation step}, the central server can collect and aggregate all local gradients and update the global model as \begin{equation} \label{eqn:federated_global_update}
    u_{t + 1} \leftarrow u_t - \eta \sum_{k = 1}^K \frac{n_k}{n} g_k. 
\end{equation} Alternatively, each local client $k$ can update its local model as $\displaystyle u_{t + 1}^k \leftarrow u_t - \eta g_k$, and then have the central server collect and aggregate all updated local models as \begin{equation} \label{eqn:federated_global_local_update}
    u_{t + 1} \leftarrow \sum_{k = 1}^K \frac{n_k}{n} u_{t + 1}^k. 
\end{equation} This algorithm is termed as \textbf{Federated SGD} in \cite{DBLP:journals/corr/McMahanMRA16}. 

In practice, averaging the neural network involves computing the average of the weight parameters between neurons based on their respective positions within the network. Specifically, if a neural network has $L + 1$ layers and $L_j$ neurons in each layer, with the first one being the input layer and the last one being the output layer, then define $\zeta_{i, j; r}^{k, t}$ as the weight parameter from the $i$-th neuron in layer $r$ to the $j$-th neuron in layer $r + 1$ with $1 \leq r \leq L$, $1 \leq i \leq L_r$, $1 \leq j \leq L_{r + 1}$, of client $k$ in iteration $t$. Then $\displaystyle u_t^k = \big\{ \zeta_{i, j; r}^{k, t}\big\}_{1 \leq r \leq L, 1 \leq i \leq L_r, 1 \leq j \leq L_{r + 1}}$. Eq.~ \eqref{eqn:federated_global_local_update} can be rewritten as 
\begin{equation} \label{eqn:federated_global_local_updata_zeta}
    \zeta_{i, j; r}^{g, t + 1} \leftarrow \sum_{k = 1}^K \frac{n_k}{n} \zeta_{i, j; r}^{k, t + 1}, \text{ for } 1 \leq r \leq L, 1 \leq i \leq L_r, 1 \leq j \leq L_{r + 1}, 
\end{equation} where $\zeta^g$ is the weight in the global model. 

Note that instead of taking one step of gradient descent on each local training step, one can add more computation to each client by iterating the local update. That is, each client can repeat $\displaystyle u_{t + 1}^k \leftarrow u_t - \eta g_k$ multiple times before uploading the model to the central server. This approach is later termed as \textbf{Federated Average}. The algorithm of federated averaging is given at Eqs.~\eqref{alg:federated_average} and \eqref{alg:ClientUpdate}. 

\begin{algorithm}
    \caption{\textsc{Federated Averaging}. \\ Run on the server. Collect and aggregate all trained local models and re-distribute the new global model to every client.} \label{alg:federated_average}
    \SetAlgoLined
    \KwData{Clients indexed by $k$; fraction of clients being selected $C$; local minibatch size $B$; number of local epochs $E$; learning rate $\eta$}
    \KwResult{Updated model $u_{i + 1}$}
    Initialize $u_0$\;
    \For{$t = 1, 2, 3, \dots$}{
        $m \leftarrow \max(C \cdot K, 1)$\;
        $S_t \leftarrow \text{randomly select } m \text{ clients}$\;
        \For{each client $k \in S_t$}{
            $u_{t + 1}^k \leftarrow \text{ClientUpdate}(k, u_t)$\;
        }
        $u_{t + 1} \leftarrow \sum_{k = 1}^m \frac{n_k}{n} u_{t + 1}^k$\;
    }
\end{algorithm}

\begin{algorithm}
    \caption{\textsc{ClientUpdate}. \\ Run on the client side. Regular optimization process done in the client side.} \label{alg:ClientUpdate}
    \SetAlgoLined
    \KwData{Local minibatch size $B$; local epochs $E$; learning rate $\eta$}
    \KwResult{Client updated model $u_{i + 1}^k$, local dataset $P_k$}
    $\mathcal{B} \leftarrow$ split $P_k$ into batches of size $B$\;
    \For{every local epoch $i = 1$ to $E$}{
        \For{$b \in \mathcal{B}$}{
            $u \leftarrow u - \eta \nabla l(b; u)$\;
        }
    }
    \KwRet{$u$}\;
\end{algorithm}

\subsection{Initialization}
In the federated learning setting, there are two ways to generate the initial models in the server and all clients: 
\begin{itemize}
    \item \textbf{Common Initialization}\\
    For common initialization methods, the server will generate the inital model $u_0$ and distribute it to each client. That is, all client will have the same model as the server before training process starts. 
    \item \textbf{Independent Initialization}\\
    For Independent initialization methods, no model will be generated by the server. Instead, every client will start with an random initial local model $u_0^k$. 
\end{itemize}

It has been shown by \cite{DBLP:journals/corr/McMahanMRA16} that the FedAvg algorithm with common initialization can successfully improve the accuracy on both training/test sets. In addition to this, \cite{DBLP:journals/corr/McMahanMRA16} also showed that the FedAvg algorithm with independent initialization will increase the training loss in general. 

Federated learning presents several significant advantages, including the preservation of privacy through the retention of data on local devices, thereby minimizing the risk of data exposure. It facilitates the efficient utilization of decentralized datasets without the need for central aggregation, while also reducing bandwidth consumption by transmitting only model updates. These attributes make federated learning particularly valuable in contexts where privacy, efficiency, and scalability are of paramount importance. 

Traditional neural networks, such as feed-forward networks and convolutional neural networks (CNNs), are typically not considered generative because they are designed primarily for discriminative tasks, such as classification and regression. These networks learn to map input data to a corresponding output by minimizing a loss function that quantifies the discrepancy between predicted and actual values. Their primary objective is to make predictions based on learned patterns from labeled data, rather than capturing the underlying distribution of input data or generating new, similar instances. Federated learning with traditional neural networks is also not regarded as a generative method, as it focuses on prediction tasks rather than modeling the underlying data distribution. While it can yield successful predictions, as demonstrated by \cite{DBLP:journals/corr/McMahanMRA16}, it does not have the capacity to capture the intrinsic distribution of the data. 

\subsection{Federated Learning with Bayesian Neural Network}
To address observation uncertainties, the Bayesian Neural Network (BNN) is employed. BNN operates under two fundamental assumptions: first, that probability quantifies the occurrence of events, and second, that prior beliefs influence subsequent posterior beliefs (\cite{BNN_intro}). In traditional neural networks, the parameters are defined as $\theta = (W, b)$ where $W$ is the weight and $b$ is the bias. The simplest architecture is given as \begin{equation}\label{eqn:traditional_NN}
    l_i = \sigma_i(W_i l_{i - 1} + b_i), 
\end{equation} where $\sigma$ is the activation function and $l_i$ is the linear transformation at the network's $i$-th layer. In contrast, BNN starts with a prior distribution over the model parametrization $p(\theta)$, and a likelihood $p(y|x, \theta)$, where $\theta$ is the model parameter, $x$ is the input and $y$ is the output. Let $D$ be the training set, $D_x$ be the training input and $D_y$ be the training output. The BNN uses Bayesian inference \begin{equation}\label{eqn:Bayesian_inference}
    p(\theta|D) = \frac{p(D_y|D_x, \theta)p(\theta)}{p(D)} = \frac{p(D_y|D_x, \theta)p(\theta)}{\int_\theta p(D_y|D_x, \theta ') p(\theta ') d\theta '}, 
\end{equation} to find the posterior $p(\theta|D)$. When using BNN for prediction, the probability distribution $p(y| x, D)$ is given as \begin{equation}\label{eqn:BNN_prediction}
    p(y|x, D) = \int_{\theta} p(y|x, \theta ') p(\theta '|D)d \theta '.
\end{equation}

Although Eq.~\eqref{eqn:BNN_prediction} is intractable in general, it can be approximated by using the Monte Carlo method. In other words, by sampling $N$ models with $\theta^{(n)} \sim p(\theta| D)$, one can rewrite Eq.~\eqref{eqn:BNN_prediction} as 
\begin{equation}\label{eqn:BNN_prediction_MC}
    p(y|x, D) \approx \frac{1}{N} \sum_{n = 1}^N p\big(y|x, \theta^{(n)}\big).
\end{equation} 

In \cite{chen2021fedbemakingbayesianmodel}, the authors proposed FedBE by applying the Bayesian model to the global model. In FedBE, the posterior $p(\theta| D)$ is assumed to have either the Gaussian or Dirichlet distribution and can be constructed by the information from local models. Then $N$ models $\theta^{(n)}$ are sampled from the posterior $p(\theta|D)$, and later used to perform ensemble for the prediction scheme \eqref{eqn:BNN_prediction_MC}. The predictions $\{p(y_i|x_i, D)\}$, along with the stochastic weight averaging(SWA) method, will be applied to update the global model. 

While Bayesian model ensembles can enhance accuracy, FedBE introduces uncertainty only in the global model. Specifically, FedBE employs Bayesian inference at the central server to determine the posterior distribution $p(\theta| D)$. However, each client trains his model alone, and the client model remains a deterministic model. The client's model cannot measure the potential noise in his local data. 

A potential approach to quantifying noise on the client side is to implement BNN for all models. In this framework, both the global model and all clients' local models would be BNNs. Each client would train a local BNN and subsequently transmit the updated posterior and predictions to the central server to optimize the global model. However, this approach is intractable in general. In BNNs, parameters such as weights, biases, or activation functions are typically treated as random variables (\cite{BNN_intro}). In the context of federated learning, the central server is responsible for aggregating client models to refine the global model. Since federated learning necessitates the aggregation of client models, incorporating local BNNs would present significant challenges, as aggregating random variables is inherently complex. On the other hand, if all parameters are scalar values or numerical data, their aggregation within the federated learning framework can be conducted with ease. Thus, in our pursuit of making federated learning generative, specifically by enabling it to capture the underlying data distribution, we introduce an alternative neural network architecture, the stochastic neural network, in the next section.

\section{Stochastic Neural Network} \label{sec:SNN}
The Stochastic Neural Network (SNN) architecture builds on the ``Neural ODE" framework, which describes the progression of hidden layers in a deep neural network (DNN) as a discretized ordinary differential equation (ODE) system. This approach offers a mathematical basis for residual neural networks, an essential structure in machine learning. To incorporate uncertainty, Gaussian noise is added to the hidden layers instead of treating parameters as random variables, as is done in the Bayesian approach. This results in a stochastic ordinary differential equation (SDE) formulation of DNNs, referred to here as the stochastic neural network (SNN). The key difference between SNNs and Bayesian neural networks (BNNs) is that SNNs derive uncertainty from the diffusion term of the SDE, with the diffusion coefficient controlling the network's probabilistic outputs. By estimating these diffusion coefficients, the SNN framework effectively enables uncertainty quantification for probabilistic learning.

\subsection{Stochastic optimal control formulation of deep probabilistic learning}
For the SNN structure, we consider the following model
\begin{equation}\label{eqn:SNN_dynamical_system}
    X_{n + 1} = X_n + hf(X_n, u_n) + \sqrt{h}g(u_n) \omega_n, 
\end{equation} where $X_n := [x_n^1, x_n2, \dots, x_n^L] \in \mathbb{R}^L$ is the vector containing $L$ neurons in the $n$-th layer of the network, $f$ is an activation function, $u_n$ is the set of network parameters in the $n$-th layer, $h$ is a positive stabilization constant, $\omega_n$ is a standard $L$-dimensional Gaussian random variable characterizing the uncertainty, and $g$ is a coefficient function measuring the size of uncertainty. The initial state $X_0$ of the dynamical system \eqref{eqn:SNN_dynamical_system} represents the input and $X_N$ as the output. We model the SNN as a stochastic sequence in the form of a discrete stochastic differential equation, where the function $f$ defines the drift term and $g$ defines the diffusion term (see \cite{archibald2021backwardsdemethoduncertainty}).

Now, if we choose a positive time $T$ as the terminal time and let $N \to \infty$ or $h \to 0$ with $h = \frac{T}{N}$, then the dynamical system \eqref{eqn:SNN_dynamical_system} can be rewritten as an integral form: \begin{equation} \label{eqn:SNN_integral_form}
    X_{T} = X_0 + \int_0^T f(X_t, u_t)dt + \int_0^T g(u_t) dW_t, 
\end{equation} where $W:=\{W_t\}_{0\leq t\leq T}$ is a standard Brownian motion corresponding to the i.i.d. Gaussian noise $\omega_n$ in Eq.~\eqref{eqn:SNN_dynamical_system}. The differential form of Eq.~\eqref{eqn:SNN_integral_form} is 
\begin{equation} \label{eqn:SNN_differential_form}
    dX_t = f(X_t, u_t)dt + g(u_t)dW_t, \hspace{1cm} 0\leq t \leq T
\end{equation} with the state process $X_t$. 

By treating $u$ as a parameter to be optimized in a learning process, the question is considered as a stochastic optimal control problem, and the goal is to find a parameter $u^*$ that minimizes the discrepancy between the SNN output and the training data. That is, define the cost function as \begin{equation}\label{eqn:SNN_cost}
    J(u):= \mathbb{E}[\Phi(X_T, \Gamma)], 
\end{equation} where $\Gamma$ is a random variable which generates training data to be compared with the SNN out, and a loss function $\Phi(X_T, \Gamma) = \| X_T - \Gamma \|_{loss}$ for some loss functions $\|\cdot \|_{loss}$ depending on the context. The goal is to find an \textit{optimal control} $u^*$ such that $$J(u^*) = \inf_{u \in \mathcal{U}[0, T]} J(u), $$ where $U[0, T]$ is an admissible control set. 

The dynamic programming and the stochastic maximum principle are two common methods in solving a stochastic optimal control problem. The dynamic programming approach addresses the stochastic optimal control problem by solving the Hamilton-Jacobi-Bellman(HJB) equation, a nonlinear partial differential equation(PDE), through numerical methods. In contrast, the stochastic maximum principle approach aims to determine the optimal control by satisfying the optimality conditions of the Hamiltonian function, typically using gradient descent optimization techniques. In machine learning, neural networks often have a large number of neurons, and this number corresponds to the dimensionality of the HJB equations. Therefore, it is expensive to apply the dynamic programming principle. 

\subsection{Stochastic Maximum Principle Solution for Stochastic Optimal Control Problem}
First, we assume that the optimal solution $u^* \in \mathcal{U}[0, T]$. To derive the optimal solution, one can find the gradient of $J$ as \begin{equation} \label{eqn:SNN_gradient_of_loss}
    \nabla J_u(u_t) = \mathbb{E} [f_u(X_t, u_t)^\top Y_t + g_u(u_t)^\top Z_t], 
\end{equation} where $J_u, f_u, g_u$ are the partial derivatives of $J, f, g$, respectively, with respect to $u$, and the pair $(Y_t, Z_t)$ is an adapted solution to an adjoint backward stochastic differential equation(BSDE) defined as: \begin{equation}\label{eqn:SNN_BSDE}
    dY_t = -f_x(X_t, u_t)^\top Y_tdt + Z_tdW_t, \hspace{1cm} Y_T = \Phi_x'(X_T, \Gamma),
\end{equation} where $f_x$ is the partial derivative of $f$ with respect to $X$ (\cite{archibald2021backwardsdemethoduncertainty}). The solution $Y$ propagates backward from $t = T$ to $t = 0$ with the initial condition listed in Eq.~\eqref{eqn:SNN_BSDE} and $Z$ is the martingale representation of $Y$ (see \cite{Bao2018, BSDE_filter}). Following the idea of gradient descent, with an initial guess $u_0$ and learning rate $\eta_k$, one can find the optimal solution as: 
\begin{equation} \label{eqn:SNN_cost_optimization}
    u_t^{k + 1} = u_t^k - \eta_k \nabla J_u(u_t^k), \hspace{1cm} k = 0, 1, 2, \dots, \hspace{0.5cm} 0 \leq t \leq T. 
\end{equation}

Note that in the classical formulation of the Stochastic Maximum Principle (SMP), the optimal control is characterized as the maximizer of a Hamiltonian function (\cite{peng1990general}). That is, given the dynamic system \eqref{eqn:SNN_integral_form} and the cost function in Eq.~\eqref{eqn:SNN_cost}, the Hamiltonian is given as: 
\begin{equation} \label{eqn:SMP_hamiltonian}
    H(t, X, u, P, Q) = f(X_t, u_t)^\top P_t + g(u_t)^\top Q_t. 
\end{equation} with the adjoint BSDE as 
\begin{equation} \label{eqn:BSDE_hamiltonian}
    dP_t = -H_x(t, X_t, u_t, P_t, Q_t) dt + Q_tdW_t, \hspace{1cm} P_T = \Phi_x'(X_T, \Gamma).
\end{equation} In SMP, it states that the optimal control $u_t^*$ maximizes the Hamiltonian: 
\begin{equation} \label{eqn:maximize_hamiltonian}
    u_t^* = \arg \max H(t, X_t, u, Y_t, Z_t), \text{ a.e., } t \in [0, T], \text{ a.s.}
\end{equation} Here, however, instead of maximizing the Hamiltonian, we calculate the gradient of the cost function $J$ with respect to the control $u$ given in Eq.~\eqref{eqn:SNN_gradient_of_loss}, which, upon comparison, is equivalent to the derivative of the Hamiltonian in the classical SMP formulation: \begin{equation} \label{eqn:gradient_hamiltonian}
    \nabla_u H = f_u(X_t, u_t)^\top Y_t + g_u(u_t)^\top Z_t. 
\end{equation} In the classical SMP method, maximizing the Hamiltonian $H$ leads to minimum cost $J(u)$, because of the negative drift term in the adjoint BSDE \eqref{eqn:BSDE_hamiltonian}. In our approach, directly minimizing the cost function $J(u)$ using the gradient $\nabla_u H$ achieves the same goal (\cite{archibald2021backwardsdemethoduncertainty}). 

\subsection{Numerical Method on Determining the Optimal Solution}
Implementing the stochastic gradient descent method requires finding the numerical approximation to the SDE \eqref{eqn:SNN_integral_form} and the BSDE \eqref{eqn:SNN_BSDE}. To solve the BSDE \eqref{eqn:SNN_BSDE}, we assume the discrete time points over the interval $[0, T]$ as $\Pi^N = \{t_n | 0 = t_0 < t_1 < t_2 < \cdots < t_N = T\}$ and $h$ as the step size of the partition. 

\subsubsection{Numerical Solution on Backward Stochastic Differential Equation}
For the forward SDE \eqref{eqn:SNN_integral_form}, we apply the Euler-Maruyama scheme and derives the approximation of the state variable $X$ on sub-interval $[t_n, t_{n + 1}]$ as 
\begin{equation} \label{eqn:numerical_SDE_X}
    X_{t_n + 1} = X_{t_n} + hf(X_{t_n}, u_{t_n}) + g(u_{t_n}) \Delta W_{t_n} + E_X^n, 
\end{equation} where $\Delta W_{t_n} = W_{t_{n + 1}} - W_{t_n}$ and $E_X^n = \int_{t_n}^{t_{n + 1}} f(X_r, u_r) dr - hf(X_{t_n}, u_{t_n}) + \int_{t_n}^{t_{n + 1}} g(u_r) dW_r - g(u_{t_n})\Delta W_{t_n}$ is the approximation error. 

For the BSDE \eqref{eqn:SNN_BSDE}, integrating both sides on interval $[t_n, t_{n + 1}]$ will give 
\begin{equation} \label{eqn:BSDE_integral_form}
    Y_{t_n} = Y_{t_{n + 1}} + \int_{t_n}^{t_{n + 1}} f_x(X_r, u_r) Y_r dr - \int_{t_n}^{t_{n + 1}} Z_r dW_r. 
\end{equation} For $Y$ is adapted with respect to $W$, then $\mathbb{E}[Y_{t_n} | X_{t_n}] = Y_{t_n}$ (see \cite{Bao_Control2020}). Now, by taking the conditional expectation of $X_{t_n}$ on Eq.~\eqref{eqn:BSDE_integral_form} and using the right point approximation to approximate the deterministic integer, one can derive \begin{equation} \label{eqn:numerical_BSDE_Y}
    Y_{t_n} = \mathbb{E}[Y_{t_{n + 1}} | X_{t_n}] + h\mathbb{E} \left[f_x \left(X_{t_{n + 1}}, u_{t_{n + 1}} \right) Y_{t_{n + 1}} | X_{t_n}\right] + E_Y^n, 
\end{equation} where $E_Y^n$ is the error term (see \cite{Bao2022_KL, Bao_first}). 

Finally, for the martingale representation $Z$, apply the left point approximation on the deterministic integral in Eq.~\eqref{eqn:BSDE_integral_form} and get 
\begin{equation} \label{eqn:numerical_BSDE_Z}
    Y_{t_n} = Y_{t_{n + 1}} + hf_x \left(X_{t_n}, u_{t_n} \right)Y_{t_n} - Z_{t_n} \Delta W_{t_n} + E_Z^n, 
\end{equation} where $E_Z^n$ is the error term in approximation (see \cite{archibald2021backwardsdemethoduncertainty}). Note that both $X$ and $Y$ are adapted to $W$. Therefore, if one multiplies $\Delta W_{t_n}$ on both sides of Eq.~\eqref{eqn:numerical_BSDE_Z} and takes the conditional expectation on $X_{t_n}$ to it, one shall have 
\begin{equation} \label{eqn:numerical_BSDE_Z_1}
    \mathbb{E} \left[ Z_{t_n} | X_{t_n}\right]h = \mathbb{E} \left[ Y_{t_{n + 1}} \Delta W_{t_n} | X_{t_n} \right] + \mathbb{E} \left[ E_Z^n \Delta W_{t_n} | X_{t_n} \right].
\end{equation} Now, by ignoring all error terms in Eqs.~\eqref{eqn:numerical_SDE_X}, \eqref{eqn:numerical_BSDE_Y}, and \eqref{eqn:numerical_BSDE_Z_1}. We adopt the numerical solution for the forward and backward SDE as: 
\begin{align}
    X_{t_n + 1} &= X_{t_n} + hf(X_{t_n}, u_{t_n}) + g(u_{t_n}) \Delta W_{t_n}, \label{eqn:numercial_SDE_XX}\\
    Y_{t_n} &= \mathbb{E}[Y_{t_{n + 1}} | X_{t_n}] + h\mathbb{E} \left[f_x \left(X_{t_{n + 1}}, u_{t_{n + 1}} \right) Y_{t_{n + 1}} | X_{t_n}\right], \label{eqn:numercial_BSDE_YY}\\
    Z_{t_n} &= \frac{\mathbb{E} \left[ Y_{t_{n + 1}} \Delta W_{t_n} | X_{t_n}\right]}{h} \label{eqn:numerical_BSDE_ZZ}. 
\end{align} 
Finally, one can use Eqs. \eqref{eqn:SNN_gradient_of_loss} and \eqref{eqn:SNN_cost_optimization} to determine the optimal control as \begin{equation}
    u_n^{k + 1} = u_n^k - \eta_k \mathbb{E} \left[ f_u \left( X_n^k, u_n^k \right)^\top Y_n^k + g_u(u_n^k)^\top Z_n^k \right]
\end{equation} in iteration $k$ where $X_n^k, Y_n^k, Z_n^k$ are numerical approximations obtained by $u_n^k$ in interval $[t_n, t_{n + 1}]$. 

\subsubsection{Monte Carlo Approximation on Conditional Expectation}
Now, the question has been changed from finding the numerical solution for a stochastic optimal control problem to numerically approximating the conditional expectation $\mathbb{E}[\cdot | X_{t_n}]$ in Eqs.~\eqref{eqn:numercial_BSDE_YY} and \eqref{eqn:numerical_BSDE_ZZ}. In \cite{archibald2021backwardsdemethoduncertainty}, we introduced the Monte Carlo simulation to approximate the expectation. To avoid the potential computation failure of the Monte Carlo method when the SNN is deep or the Monte Carlo sampling number needs to be very large with high-dimensional state $X$, we follow the idea of the Stochastic Gradient Descent (SGD). That is, instead of using the mean of all Monte Carlo samples for expectation, we can choose only one data sample to approximate the expectation at each iteration. Specifically, at iteration $k$ with the control $u_n^k$, a Gaussian sample $\omega_n \sim N(0, h)$($h$ is the variance) is generated, and we evaluate the conditional expectation $\mathbb{E}[\cdot | X_{t_n}]$ only at this sample point. Therefore, we introduce the following sample-wise numerical solutions $X_n^k$ of $X$, $(Y_n^k, Z_n^k)$ of $(Y, Z)$ as 
\begin{align}
    X_{n + 1}^k &= X_n^k + hf(X_n^k, u_n^k) + g(u_n^k)\omega_n^k, \label{eqn:numerical_SDE_XXX}, \\
    Y_n^k &= Y_{n + 1}^k + h f_x(X_{n + 1}^k, u_{n + 1}^k)^\top Y_{n + 1}^k \label{eqn:numerical_BSDE_YYY}, \\
    Z_n^k &= \frac{Y_{n + 1}^k \omega_n^k}{h} \label{eqn:numerical_BSDE_ZZZ}, 
\end{align} 
where $\omega_n^k$ is the $k$-th realization of $\omega_n$ (\cite{SNN_Convergence}). Notice that in both Eq.~\eqref{eqn:numercial_BSDE_YY} and Eq.~\eqref{eqn:numerical_BSDE_ZZ}, the expectations are now represented by a single realization of samples indexed by $k$. Then, at each iteration step, we generate a sample path $\{X_n^k\}_n$ for the state and solve the BSDE with this sample path using Eq.~\eqref{eqn:numerical_BSDE_YYY} and Eq.~\eqref{eqn:numerical_BSDE_ZZZ}. Now, one can approximate the gradient of the cost function $J$ as 
\begin{equation}\label{eqn:numerical_cost_final}
    \nabla J_u^k(u_n^k) = f_u(X_n^k, u_n^k)^\top Y_n^k + g_u(u_n^k)^\top Z_n^k, 
\end{equation} and the gradient descent process as 
\begin{equation} \label{eqn:numerical_gradient_descent_final}
    u_n^{k + 1} = u_n^k - \eta_k \nabla J_u^k(u_n^k). 
\end{equation}

It should be mentioned that since only one sample $X$ is used in the Monte Carlo approximation and the simulated pair $(Y, Z)$ only describes the solution of the adjoint BSDE corresponding to the given simulated path $\{X_n^k\}_n$, the numerical approximation for $Y$ and $Z$ are incomplete. However, we need to point out that the purpose of solving the optimal control problem is to find the optimal control $u^*$, or the optimal parameters in SNN, rather than to obtain an accurate numerical solution for $(Y, Z)$. 

One can find the algorithm for solving the SNN model at Algorithm \ref{alg:SNN}. 

\begin{algorithm}
    \caption{\textsc{SMP approach for SNN} \\ Find the optimal control $u_n^*$ by using the SMP.}\label{alg:SNN}
    \KwData{A SNN model as equation \ref{eqn:SNN_dynamical_system}; a cost function as equation \ref{eqn:SNN_cost}; a partition $\Pi^N$; the number of iteration steps $K \in \mathbb{N}$; the learning rate $\{\eta_k\}_K$; the initial guess $u_n^0$. }
    \KwResult{The optimal control $u_n^*$. }
    \For{Stochastic gradient descent step $k = 0, 1, \dots, K - 1$}{
        Use equation \ref{eqn:numerical_SDE_XXX} to simulate a realization of the state process $X_n^k$\;
        Use equation \ref{eqn:numerical_BSDE_YYY} and equation \ref{eqn:numerical_BSDE_ZZZ} to simulate a solution pair $(Y_n^k, Z_n^k)$ of the adjoint BSDE equation\;
        Use equation \ref{eqn:numerical_cost_final} to calculate the gradient of the cost function $J$ and use equation \ref{eqn:numerical_gradient_descent_final} to find the updated control $u_n^{k + 1}$\;
    }
    The optimal control is given by $u_n^* = u_n^{K}$\;
    \KwRet{$u_n^*$}\;
\end{algorithm}

The convergence analysis has been studied in \cite{SNN_Convergence, Liang_Control,Sun_Control} and one can find more numerical experiments in \cite{archibald2021backwardsdemethoduncertainty}. Since SNN is designed to numerically solve SDEs, it is capable of not only finding the true data states, but also capturing the latent uncertainty in the data. Since all parameters (neural networks' weights and biases) are deterministic in SNN, we consider it more suitable for federated learning, since aggregating random variables in BNN is more challenging than numbers in SNN. The federated learning method with SNN as the global and clients' models is elaborated in the next section. 

\section{Federated SNN} \label{sec:FedSNN}
We follow the idea of federated learning to construct the  federated SNN(FedStNN) method. Consider a group of $K$ clients and a training dataset $\overline{\Gamma}$ which contains all possible true observation data (no errors). Let $\{\Gamma_k\}_{k = 1}^K \displaystyle$ be a sequence of subsets of $\overline{\Gamma}$. Let $n_k = |\Gamma_k|$ represent the cardinality of dataset $\Gamma_k$ and \begin{equation}\label{eqn:FedSNN_cardinality_all_training_data}
    n = \left| \bigcup_{k = 1}^K \Gamma_k \right|. 
\end{equation} Notice that $\displaystyle \bigcup_{k = 1}^K \Gamma_k \subseteq \overline{\Gamma}$. The union of all subsets $\Gamma_k$ might not be the entire possible observation dataset $\overline{\Gamma}$, indicating that some possible observation data might not be included (observed) in any subset $\Gamma_k$. 

In federated learning, data is exclusively gathered by clients and is never shared with the central server. However, measurement noise can occur due to limitations in instruments, human errors—such as mistakes in manual data collection or interpretation—and external factors like temperature fluctuations and humidity. These challenges can prevent individuals from accurately assessing the true state of the model $\overline{\Gamma}$. Therefore, to account for the observation errors, we include a random noise $E_k$ to each subset $\Gamma_k$. Then, for all $K$ clients, their local datasets $\{P_k\}_{k = 1}^K \displaystyle$ are given as \begin{equation}\label{eqn:FedSNN_local_dataset}
    P_k = \Gamma_k + E_k, 
\end{equation} where $E_k \sim N(0, \sigma_k^2)$ is a Gaussian noise with mean $0$ and variance $\sigma_k^2$. 

It can be seen that federated learning, specifically the FedAvg Algorithm \ref{alg:federated_average}, is a deterministic machine learning model. The global model $\omega_{t}$ is updated by Eq.~\eqref{eqn:federated_global_local_update}, which is the weighted average of all updated client models. However, local clients in federated learning use the gradient descent method to optimize their models by calculating the gradients of their local data through Eq.~\eqref{eqn:federated_gradient}. Although the initial model $u_0$ distributed to each client at the beginning of the process can be generated randomly, and each client may use the stochastic gradient descent(SGD) method instead of the gradient descent(GD) method to optimize his local model $u_t^k$, the FedAvg algorithm is, in general, a deterministic model and can barely capture the randomness hiding in the training data. Since it is deterministic, federated learning with deterministic local models cannot be generative. 

On the other hand, the SNN structure shown in the previous section is a stochastic machine learning method. It carries a Gaussian random variable $\omega_n$ to measure the uncertainty of the model in its governing Eq.~\eqref{eqn:SNN_dynamical_system}. Therefore, the SNN model is more suitable for dealing with perturbed training data $\Gamma = \overline{\Gamma} + E$, where $\overline{\Gamma}$ is the real model parameter and $E$ stands for the observation error. It has been presented in \cite{archibald2021backwardsdemethoduncertainty} that the SNN model can not only determine the real model parameter in the system, but also give an error band, which can be used in generative learning. 

Since federated learning has shown a significant advancement in privacy protection by reducing the need for raw data transmission and enabling decentralized model training, and the SNN model has the capability to recognize and interpret the inherent randomness present within the training data, we are going to define a new model called \textbf{federated SNN}(\textbf{FedStNN}) which is a combined model of federated learning and the SNN structure. It will keep the local data unshareable to protect the privacy of every client, and use the idea of SNN to capture the stochasticity. 

Then, instead of using a traditional neural network model, a SNN model with parameter $u_0$ is randomly generated at the central server and then distributed to every client. For each iteration step $i$, a fraction of $CK$ clients with $0 < C \leq 1$ are selected. Then all selected clients will train their local SNN model $u_i^k$ following the stochastic neural network Algorithm \ref{alg:SNN}, and return the trained local SNN model $u_{i + 1}^k$ back to the central server. 

Specifically, for the SNN structure, there are two parallel neural networks, where one is the drift network and the other is the diffusion network. The drift network is similar to a regular neural network representing the drift term $f$ in the SNN model \eqref{eqn:SNN_dynamical_system}. The diffusion network is used to measure the randomness and the diffusion term $g$ in this model. That is, $u = (u_\alpha, u_\beta)$ where $u_\alpha$ is the drift network and $u_\beta$ is the diffusion network. 

When all selected clients have returned the trained local model $\displaystyle \{u_{i + 1}^k\}_{k = 1}^{CK}$ back to the server, the server will aggregate the drift and the diffusion network separately. That is, let $m = \displaystyle \sum_{k = 1}^{CK} n_k$, 
\begin{equation} \label{eqn:FedSNN_global_update_final}
    u_{i + 1} = (u_{i + 1, \alpha}, u_{i + 1, \beta}) = \left( \sum_{k = 1}^{CK} \frac{n_k}{m} u_{i + 1, \alpha}^k, \sum_{k = 1}^{CK} \frac{n_k}{m} u_{i + 1, \beta}^k \right). 
\end{equation} Once the server obtains the updated global model $u_{i + 1}$, it will redistribute the model $u_{i + 1}$ to every client, and the process will repeat. One can find the algorithm of FedStNN in Algorithm \ref{alg:FedSNN}. 

Within our method's framework, we adopt the concept of federated learning, where parameters are shared instead of raw data. This approach safeguards the privacy of each client. Additionally, leveraging the SNN’s ability to capture randomness in the training dataset, each client submits a local model $u_{i + 1}^k$ that not only reflects the drift term but also accounts for local random noise in Eq.~\eqref{eqn:SNN_dynamical_system}. In the experiments in the following section, we will demonstrate that the updated global model $u_{i + 1}$ derived from all local models by Eq.~\eqref{eqn:FedSNN_global_update_final} can effectively measure noise across the entire training dataset as well. 

Since all data in federated learning is collected by clients, local datasets may contain noise errors $E_K$. We have previously discussed how the SNN model can be utilized to quantify this randomness. Beyond errors, a local dataset may also exhibit significant bias due to the diverse backgrounds among clients. A client with a biased local dataset may obtain a trained local model $u_{i + 1}^k$ that can only work for a small portion of but not the whole possible training dataset. We term the case where all local datasets are IID and are representative of the entire possible training dataset as the IID case, and term the case where all local datasets are non-IID and no clients' local dataset can represent the whole population as the non-IID case. In the next experiments section, we will show that the FedStNN method can not only capture the general pattern of the whole dataset, but also measure the random noise in the dataset, even for the non-IID cases. 

\begin{algorithm}
    \caption{\textbf{FedStNN} Find the optimal model parameter $u^*$ of the stochastic neural network under the federated learning setting.} \label{alg:FedSNN}
    \KwData{A group of $K$ clients; client local dataset $\{P_k\}_k{k = 1}^K$; the fraction of clients being selected $C$; the learning rate $\eta$. }
    \KwResult{Optimal model parameter $u^*$. }
    Randomly generate the initial SNN model $u_0$ at the central server and distribute it to every client\;
    \For{$i = 1, 2, 3, \dots, N$}{
        $m \leftarrow \max(C\cdot K, 1)$\;
        randomly select $m$ clients\;
        \For{every selected client $k$}{
            use the local dataset $P_k$ and algorithm \ref{alg:SNN} where the current local model $u_i$ is the initial guess $u_n^0$ in algorithm \ref{alg:SNN} and $\eta$ as the learning rate to train and find the updated local control $u_{i + 1}^k$\;
            upload the updated local control $u_{i + 1}^k$ to the central server\;
        }
        collect all updated local optimal control $\{u_{i + 1}^k\}_{k = 1}^m$ and calculate the updated global model $u_{i + 1}$ using equation \ref{eqn:federated_global_local_update}\;
        distribute the updated global model $u_{i + 1}$ to every client\;
    }
    The optimal model parameter $u^* \leftarrow u_{N}$\;
    \KwRet{$u^*$}\;
\end{algorithm}

\section{Experiments} \label{sec:experiments}
\subsection{1D Function Approximation} \label{exp:1D}
\begin{figure}[h]
  \centering
  \begin{minipage}[b]{0.45\textwidth}
    \includegraphics[width=\textwidth]{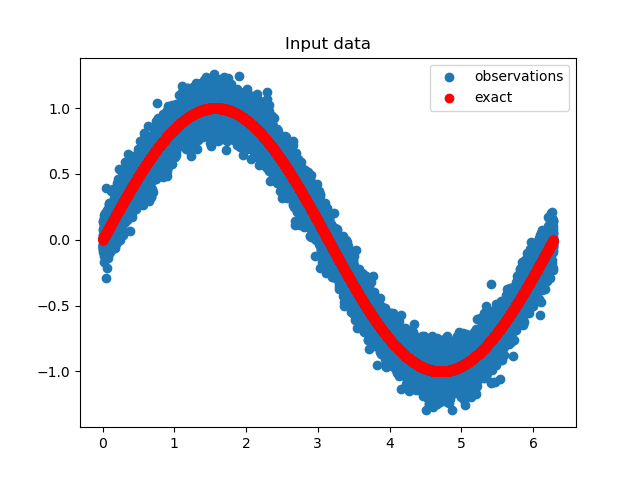}
    \caption{Example of noisy data}
    \label{fig:1D_input}
  \end{minipage}
  \hfill
  \begin{minipage}[b]{0.45\textwidth}
    \includegraphics[width=\textwidth]{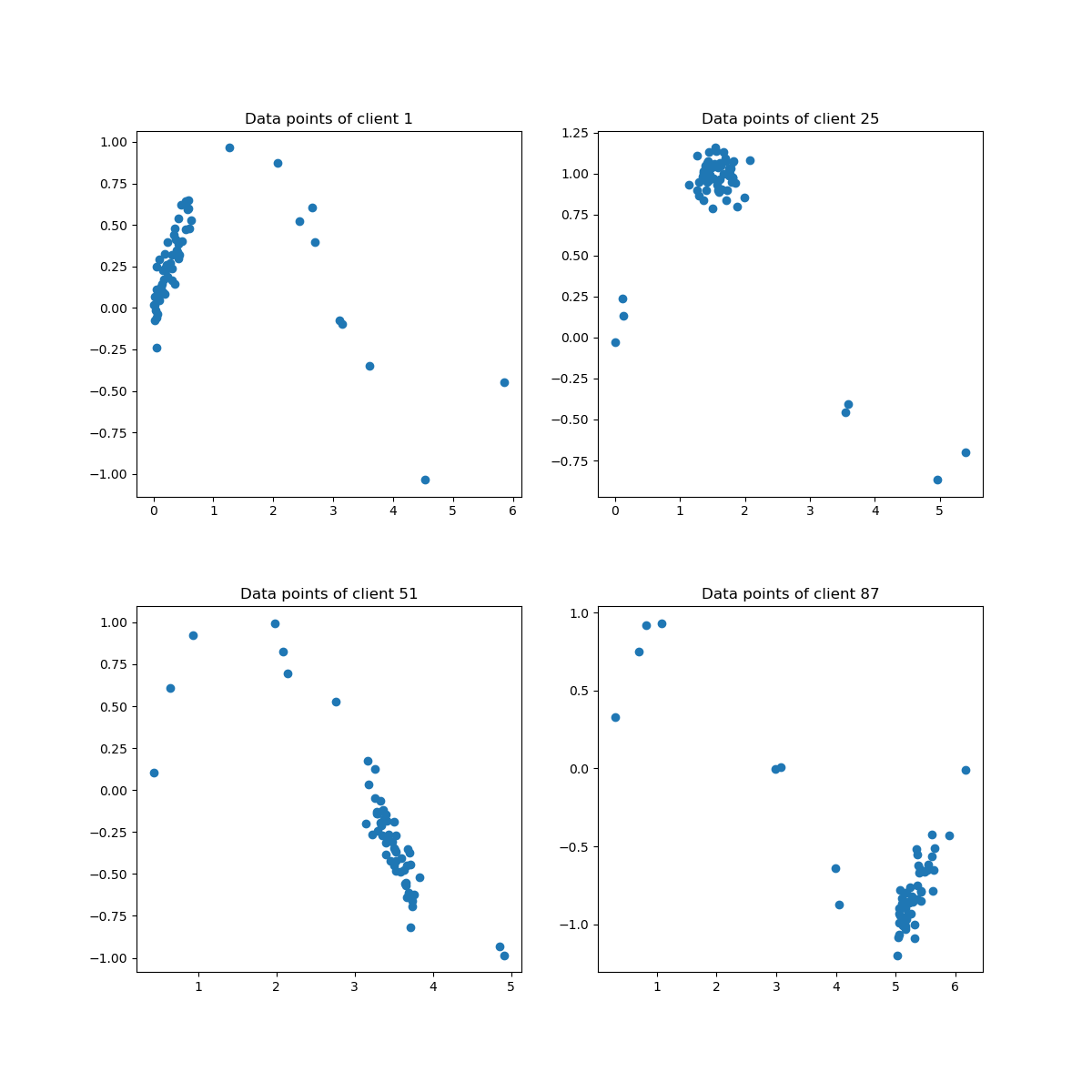}
    \caption{Examples of client local data}
    \label{fig:client_local_data}
  \end{minipage}
\end{figure}

Consider a function $f(x) = \sin(x)$ on the interval $x \in [0, 2\pi]$. We first partition the interval into 10,000 sub-intervals $[x_i, x_{i + 1}]$ with $x_1 = 0$ and $x_{10001} = 2\pi$. The function value $f(x)$ is treated as the exact value. On each point $x_i$, a random Gaussian noise with mean $\mu = 0$ and variance $\sigma^2 = 0.1^2$ is added to $f(x)$. That is, $y_i = f(x_i) + N(0, 0.1^2)$. One can find the graph of all training data in Fig. \ref{fig:1D_input} in which the exact function value $f(x)$ is shown as the red line and all noisy observations are shown as blue dots. We take $y_i$ as the observation data and distribute them to each client in two ways, IID and Non-IID. 

In federated setting, the number of clients is set to be $K = 100$ and the fraction of clients selected at each round is set to be $C = 0.1$. The local SNN model contains $4$ residual blocks with a $8\times 16 \times 8$ network in each block. Two cases of the local datasets are considered here: \begin{itemize}
    \item \textbf{IID}\\ Every client will randomly select the same amount of data points $(x_i, y_i)$ from the entire domain. That is, each client will be a representative of the population. 
    \item \textbf{Non-IID}\\ All clients will be split into 10 groups, with 10 clients in each group. Moreover, all data points will also be split into 10 groups in order, with 1,000 points in each group. All clients will still randomly select data points, but clients in group $m$ will mainly select points from data group $m$, with a few data points from other data groups. For example, client 31 through client 40 will be in client group 4 and their local dataset will mainly contain data points in between $x_{3001} = 0.6 \pi$ through $x_{4000} = 0.8\pi$, and a few points in other interval. A few clients' local dataset are shown as examples in Fig. \ref{fig:client_local_data}. Notice that for client 1, most observation data in his local dataset concentrates in the interval $[0, 0.2\pi]$, whereas for client 25, most observation data in his local dataset centers in the interval $[0.4\pi, 0.6\pi]$. 
\end{itemize}

\begin{figure}[h]
  \centering
  \begin{minipage}[b]{0.45\textwidth}
    \includegraphics[width=\textwidth]{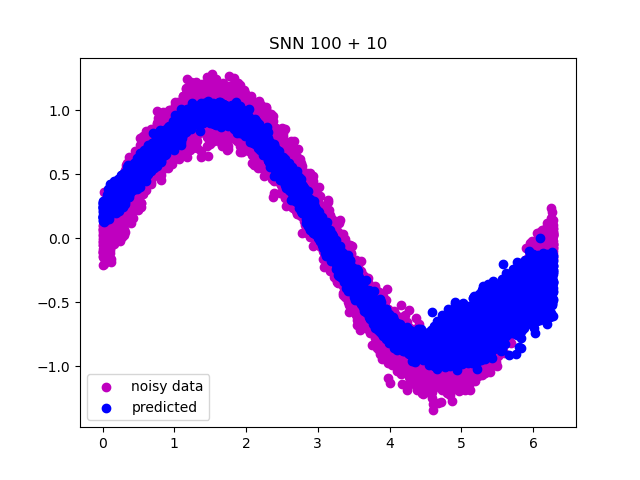}
    \caption{Global model prediction with 110 local data points}
    \label{fig:1D_non_IID_110}
  \end{minipage}
  \hfill
  \begin{minipage}[b]{0.45\textwidth}
    \includegraphics[width=\textwidth]{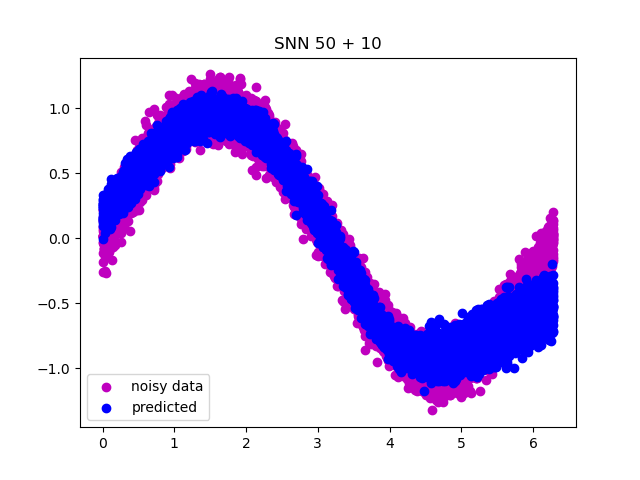}
    \caption{Global model prediction with 60 local data points}
    \label{fig:1D_non_IID_60}
  \end{minipage}
\end{figure}

For \textbf{Non-IID} case, we have clients in group $m$ select 100/50 points from data group $m$ and 10 points from other groups. After 10 federated rounds, the prediction of all points $(x_i)$ in between $[0, 2\pi]$ from the global model with 110 local training data is shown in Fig. \ref{fig:1D_non_IID_110}; the prediction of all points $(x_i)$ in between $[0, 2\pi]$ from the global model with 60 local training data is shown in Fig. \ref{fig:1D_non_IID_60}. 

In both \ref{fig:1D_non_IID_110} and \ref{fig:1D_non_IID_60}, we use purple dots to represent the observation data $y_i = f(x_i) + N(0, 0.1^2)$ and blue dots to represent the predicted value by our global model. It is clear that the global model can capture not only the drift function $f(x) = \sin(x)$, but also the Gaussian noise $N(0, 0.1^2)$ added on the function value. In Fig. \ref{fig:1D_non_IID_60}, the predicted plot almost has the same ``bandwidth'' as the observation data. In both experiments, each client has access only to their own biased local dataset, which represents a limited subset of the overall population. The following Fig. \ref{fig:single_client_prediction} illustrates the predictions of a randomly selected local model. It is evident that none of the local models can accurately predict the entire population. Nevertheless, the global model generated by our method successfully achieves accurate predictions across the entire population as shown in Fig. \ref{fig:1D_non_IID_110} and Fig. \ref{fig:1D_non_IID_60}. 

\begin{figure}
    \centering
    \includegraphics[width=0.5\linewidth]{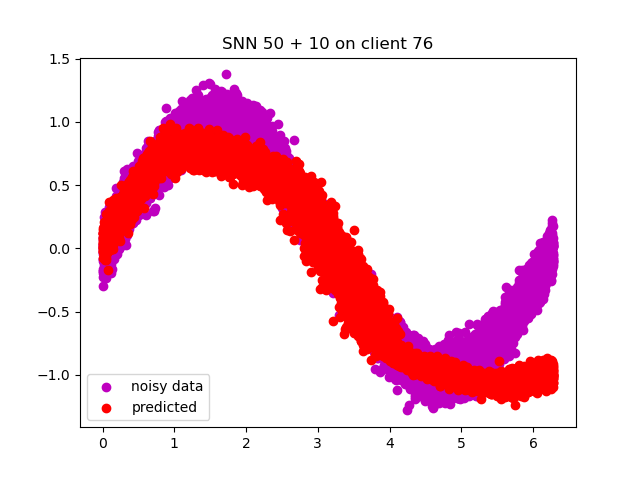}
    \caption{Prediction by client 76's local model}
    \label{fig:single_client_prediction}
\end{figure}

\subsection{2D Function Approximation} \label{exp:2d}
In this example, we will be focusing on a two-dimensional piecewise function. Let $f(x, y)$ be a piecewise function defined on the domain $[-1, 1] \times [-1, 1]$ with 
\begin{equation} \label{eqn:2D_piecewise}
    f(x, y) = \begin{cases}
x^2 + y^2, & \text{if } (x, y) \in [0, 1] \times [0, 1],\\
\sin(2\pi x) + y^2,  & \text{if } (x, y) \in [-1, 0) \times [0, 1],\\
\sin(2\pi x) + \cos(2\pi y) - 1, & \text{if } (x, y) \in [-1, 0) \times [-1, 0), \\
\cos(2\pi y) + x^2 - 1,  & \text{if } (x, y) \in [0, 1] \times [-1, 0). 
\end{cases}
\end{equation}

\begin{figure}[h]
  \centering
  \begin{minipage}[h]{0.45\textwidth}
    \includegraphics[width=\textwidth, height = 6cm]{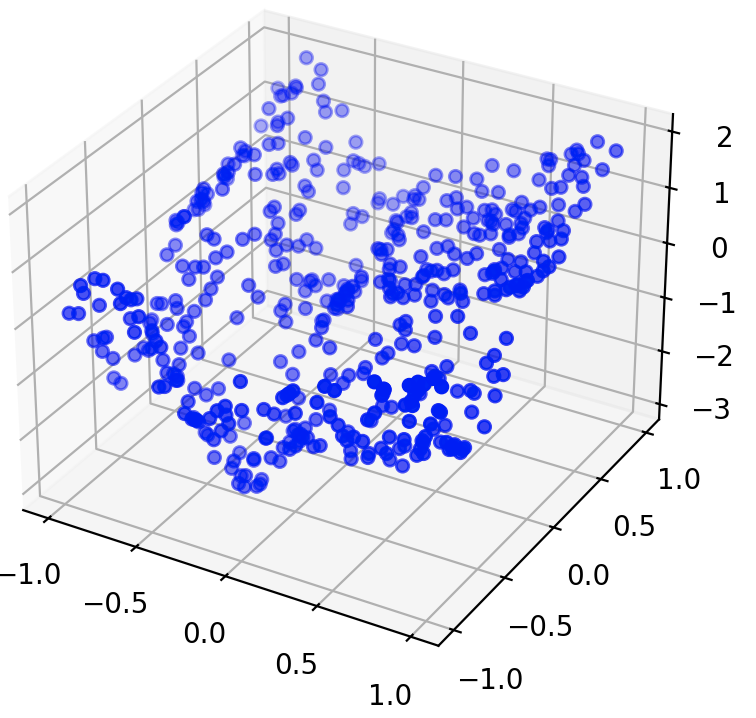}
    \caption{2D Noisy Observations}
    \label{fig:2D_noisy_data}
  \end{minipage}
  \hfill
  \begin{minipage}[h]{0.45\textwidth}
    \includegraphics[width=\textwidth, height = 6cm]{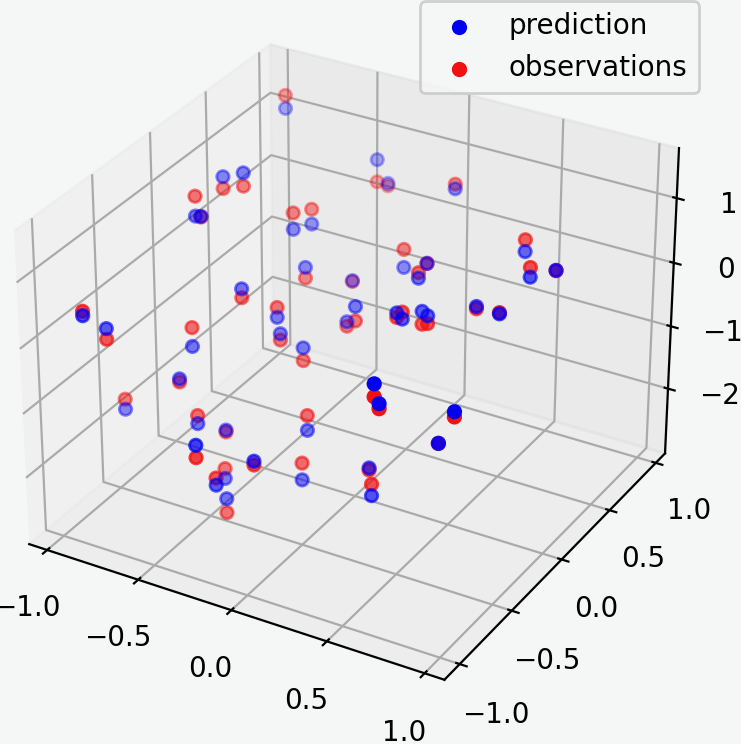}
    \caption{3D View of Prediction}
    \label{fig:2D_prediction}
  \end{minipage}
\end{figure}

Using the similar idea to finite difference methods, the $x$-axis is discretized into $101$ points with $x_0 = -1$, $x_i = -1 + 0.02i$, and $x_{100} = 1$ for $0\leq i\leq 100$; the $y$-axis is discretized into $101$ points as well with $y_0 = -1$ $y_j = -1 + 0.02j$, and $y_{100} = 1$ for $0 \leq j \leq 100$. The value $z = f(x, y) + N(0, 0.1^2)$ is the noisy observation at the input point $(x_i, y_j)$, for $0 \leq i, j\leq 100$. Graph of all noisy observation data $\{z_i\}_{i = 0}^{101^2}$ is shown at Fig. \ref{fig:2D_noisy_data}. 

In this example, only the non-IID case is considered. We assumed there to be $4$ client groups, $G_1, G_2, G_3, G_4$, and $10$ clients in each group. Every client in each client group $G_i$ will mainly focus on one quadrant of the $xy$-plane. That is, all $10$ clients in group $G_1$ will randomly choose $100$ input point pairs $(x_i, y_j) \in [0, 1] \times [0, 1]$ with the observed function values $z = f(x_i, y_j) + N(0, 0.1^2)$ to put in their own local datasets. In addition to data from quadrant $1$, each client in group $G_1$ will also randomly pick $10$ points in total from the other three quadrants. Clients in group $G_2, G_3$, and $G_4$ will choose $100$ points from quadrant II, III, IV, respectively, and $10$ additional points from the other three quadrants for their local datasets. The local SNN model has $4$ residual blocks with a $16 \times 32 \times 16$ network in each block. 

In Fig. \ref{fig:2D_prediction}, it shows $50$ randomly selected observations and the predicted values using the global model after $25$ federated rounds in a 3D plot. The MSE is approximately $2.13$. 

To give a better visualization of the difference between the real function values $f(x, y)$, the observation $z$, and the model prediction, we fixed one of the variables, either $x$ or $y$, and drew a 2D graph that includes $f(x, y), z$, and the model prediction. 

\begin{figure}
    \centering
    \includegraphics[width= 0.9\linewidth]{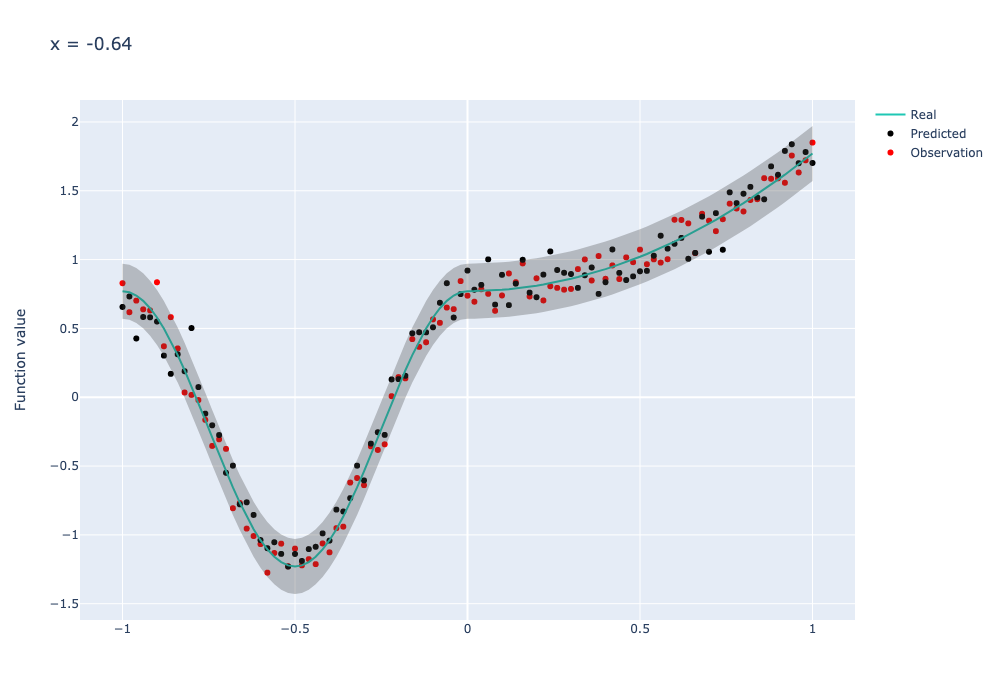}
    \caption{2D View with x = -0.64}
    \label{fig:2D_fixed_x}
\end{figure}

In Fig. \ref{fig:2D_fixed_x}, we fixed $x = -0.64$ and focused on the change in $y$. The horizontal axis shows the value of $y$. The green solid line represents the real function values $f(x, y) = f(-0.64, y)$, with the shaded band representing the function values in between two standard deviations of the Gaussian noise $N(0, 0.1^2)$. That is, it represents the area in between $f(-0.64, y) + 0.2$ and $f(-0.64, y) - 0.2$. The red dots are the observation data $z_i = f(-0.64, y_i) + N(0, 0.1^2)$ and the black dots are the global model predictions. It is clear that the global model prediction successfully captured the drift function $f(x, y)$. Moreover, instead of falling right on the green line which represents the exact function values $f(x, y)$, the black dots fill in the shaded area. 

\begin{figure}
    \centering
    \includegraphics[width=0.9\linewidth]{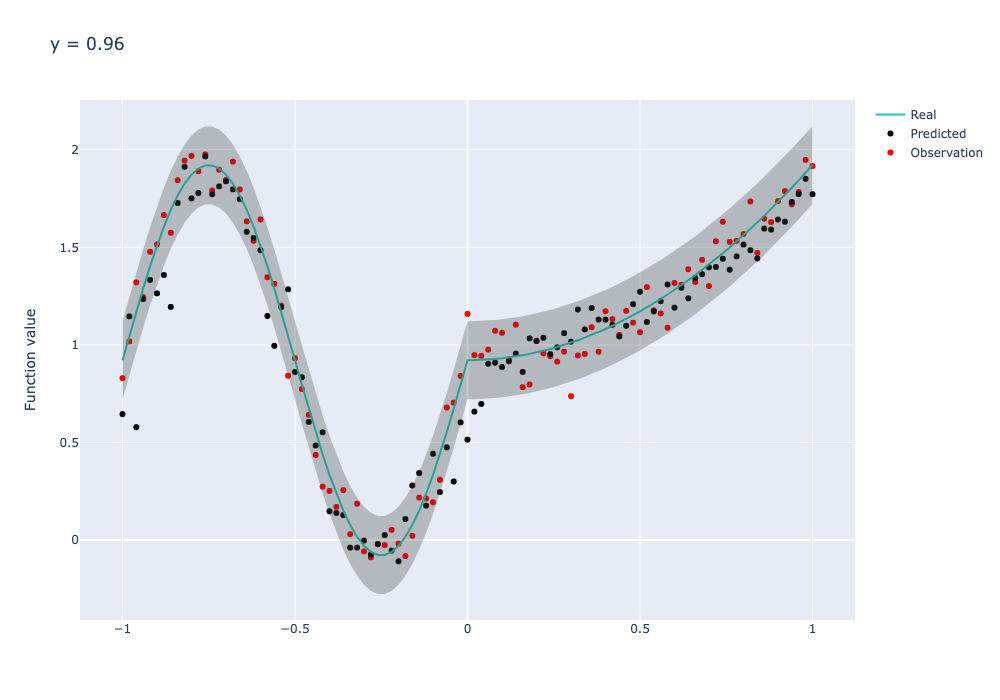}
    \caption{2D View with y = 0.96}
    \label{fig:2D_fixed_y}
\end{figure}

Similarly, in Fig. \ref{fig:2D_fixed_y}, we fixed $y = 0.96$ and focused on the change in $x$. The red dots are the observation data $z_i$ and the black dots are the predicted values by the global model at $(x, y)$ for $x \in [-1, 1]$ and $y = 0.96$. The green solid line shows the exact function value $f(x, y) = f(x, 0.96)$ with the shaded area representing the values in between $f(x, 0.96) \pm 0.2$. 

As shown in the Fig. \ref{fig:2D_fixed_x} and Fig. \ref{fig:2D_fixed_y}, all predicted values are clustered around the solid green line, demonstrating that the global model effectively captures the true function value $f(x, y)$(excluding noise). Moreover, the predicted values do not fall precisely onto the true function line but rather mostly concentrate within a two-standard-deviation band of the observation error. Since the SNN model has a drift network to capture the exact function value and a diffusion network to measure the noise, the global model can later be used to generate more observation data with a similar size of noise. 

Additionally, in this experiment, the function $f(x, y)$ is a piecewise function, and every client in each client group $G_i$ obtains more data from the corresponding quadrant than from the other three quadrants. A local model can successfully predict the function value in the quadrant with the most observation data gathered, and, by applying the FedStNN method, multiple clients can collaboratively find the suitable model parameters for the entire domain without sharing any data with each other. 

\subsection{2D Image Learning} \label{exp:2d_figure}

\begin{figure}[h]
  \centering
  \begin{minipage}[h]{0.45\textwidth}
    \includegraphics[width=\textwidth]{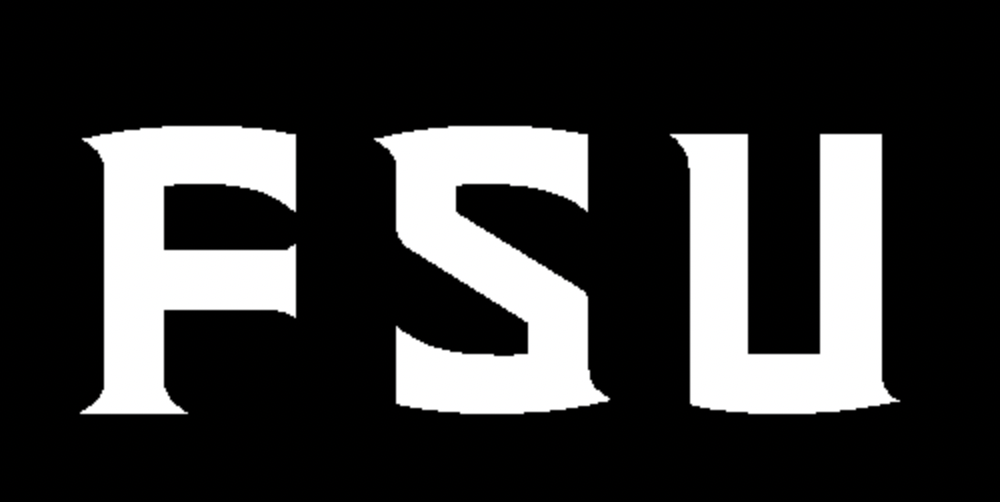}
    \caption{FSU Image}
    \label{fig:FSU}
  \end{minipage}
  \hfill
  \begin{minipage}[h]{0.15\textwidth}
    \includegraphics[width=\textwidth]{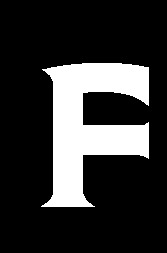}
    \caption{First sub-image}
    \label{fig:F}
  \end{minipage}
  \hfill
  \begin{minipage}[h]{0.15\textwidth}
    \includegraphics[width=\textwidth]{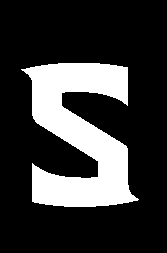}
    \caption{Second sub-image}
    \label{fig:S}
  \end{minipage}
  \hfill
  \begin{minipage}[h]{0.15\textwidth}
    \includegraphics[width=\textwidth]{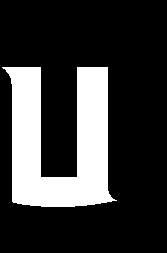}
    \caption{Third sub-image}
    \label{fig:U}
  \end{minipage}
\end{figure}

In this example, we will use the federated SNN method to learn a 2D image of letters shown in Fig. \ref{fig:FSU} with non-IID local datasets. To deal with image data, we will consider the relationship of information contained in each pixel of the image and the pixel position. That is, we will first treat the entire image as an $M \times N$ matrix where $M$ is the height and $N$ is the width of the image, measured in the number of pixels. Each entry of the matrix contains the color of the corresponding pixel. For the Fig. \ref{fig:FSU} is a black and white image, we define $C_{i, j} = 1$ for white pixel and $C_{i, j} = 0$ for black pixel with $1 \leq i \leq M$, $1 \leq j\leq N$. Then we convert the entire image matrix into a set of tuples $\displaystyle \{z_{i, j}\}_{i, j = 1}^{M, N}$ with $\displaystyle z_{i, j} = \big( (i, j); C_{i, j} \big)$. That is, the tuple of a pixel's position in the matrix and its color. The set of tuples is considered as all possible training datasets in this experiment. 

Since we intend to apply FedStNN with a non-IID case, we split the image into three subsections evenly where each letter is only contained in one subsection (in Fig. \ref{fig:F}, Fig. \ref{fig:S}, and Fig. \ref{fig:U}). Then, we assume there to be $3$ client groups $G_1, G_2, G_3$ with $10$ clients in each group. Clients in group $G_1, G_2, G_3$ will main focus on pixel information from letter F, S, U, respectively. Specifically, we let all clients randomly select $L$ pixels($z_{i, j}$) from the letter they focus on and $F$ pixels in total from the other two letters with a ratio $L:F = 100: 3$. Each client is given a SNN model containing $4$ residual blocks with a $16 \times 32 \times 16$ network in each block. The SNN model will take a pixel's position, the $(i, j)$ pair in $z_{i, j}$, as the input and trying to predict the color of that pixel $C_{i, j}$ in $z_{i, j}$. The goal is to find an SNN parameter $u^*$ for the global model such that 
\begin{equation}
    u^* = \text{argmin}_u \sum_{i, j}^{M, N} \| \Psi\Big(S\big(u; (i, j)\big)\Big) - C_{i, j} \|^2, 
\end{equation} where $\displaystyle S\big(u; (i, j)\big)$ is the global model prediction with model parameter $u$ on pixel at position $(i, j)$. Note that the pixel color is either black or white. That is, the value of $C_{i, j}$ is binary, either $0$ or $1$. However, the result from the SNN model here is numerical and unbounded. Therefore, in order to determine the color of a pixel, a binary value, based on a numerical SNN model result, we introduce the $\Psi$ function. In image \ref{fig:FSU}, there are $MN$ pixels in total with $\Xi$ white pixels and $MN - \Xi$ black pixels. After feeding all possible position pairs $(i, j)$ to the SNN model with parameter $u$, we will have a set of all numerical results $\displaystyle \left\{S\big(u; (i, j)\big) \right\}_{i, j = 1}^{M, N}$. The $\Psi$ function will map the largest $\Xi$ values in $\displaystyle \left\{S\big(u; (i, j)\big) \right\}_{i, j = 1}^{M, N}$ to $1$ and the rest to $0$.

\begin{figure}[!h]
    \centering
    \includegraphics[width=0.5\linewidth]{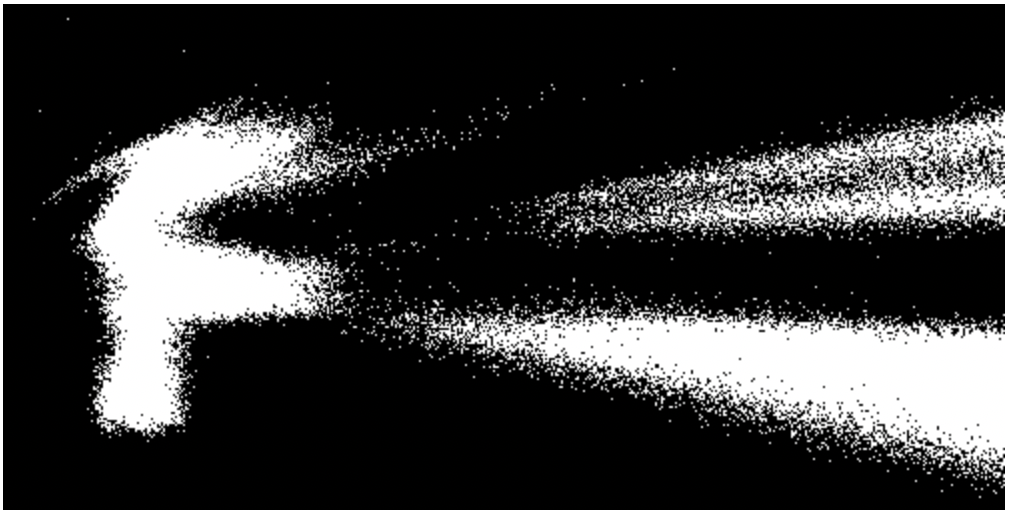}
    \caption{Client 1's Global prediction of the image}
    \label{fig:Image_single_F}
\end{figure}

\begin{figure}[!h]
    \centering
    \includegraphics[width=0.5\linewidth]{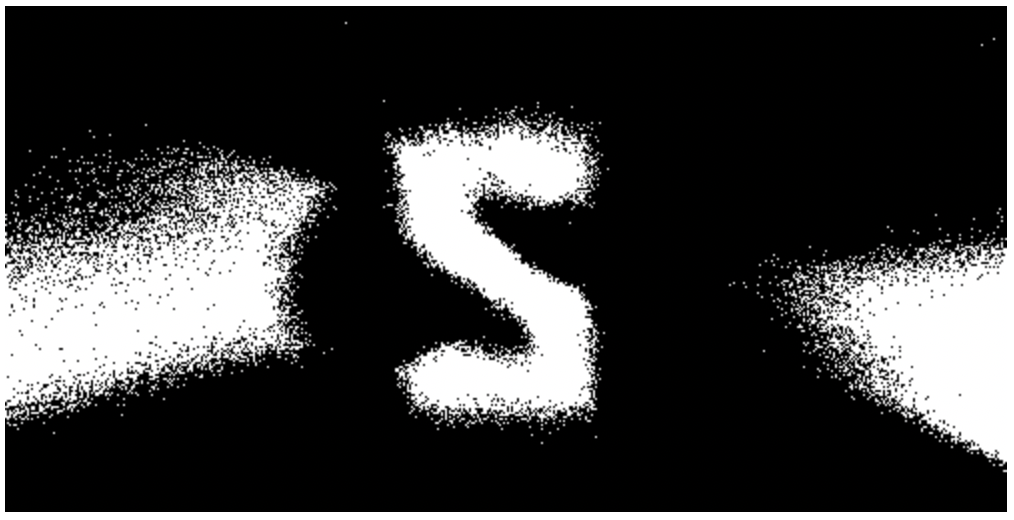}
    \caption{Client 11's Global prediction of the image}
    \label{fig:Image_single_S}
\end{figure}

\begin{figure}[!h]
    \centering
    \includegraphics[width=0.5\linewidth]{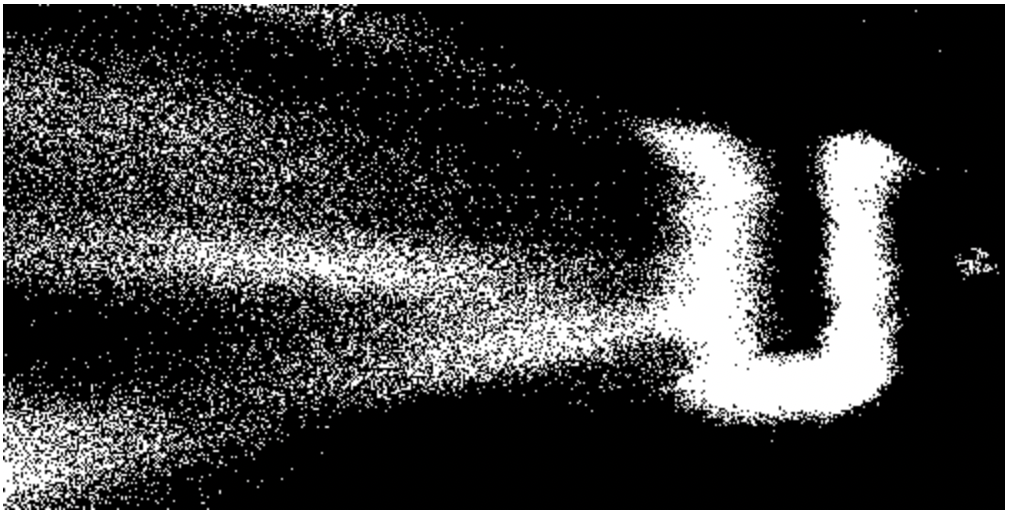}
    \caption{Client 21's Global prediction of the image}
    \label{fig:Image_single_U}
\end{figure}

Without using federated techniques, a client can only be focusing on one single letter, and the majority of that client's local data will be the pixel information from the letter he focuses on. To provide a better explanation, we manually select client 1, 11, and 21, who focus on the letters F, S, and U, respectively. Their global predictions of the image are shown in Fig. \ref{fig:Image_single_F}, \ref{fig:Image_single_S}, and \ref{fig:Image_single_U}. 

It is clear that each selected client can successfully predict the focusing letter (F for client 1, S for client 11, and U for client 21). The result can show the feasibility of the SNN model on images. However, since only a few pixel information of non-focusing letters is added to clients' local dataset, no client can predict more than one letter without information sharing. 

\begin{figure}[!h]
    \centering
    \includegraphics[width=0.5\linewidth]{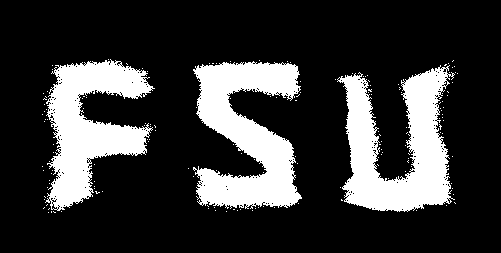}
    \caption{Global prediction of the image}
    \label{fig:Image_20000_600}
\end{figure}

Finally, the result of the global model prediction is shown in Fig. \ref{fig:Image_20000_600}. Although all clients have a few data from only one letter and limited access to the other two letters, all three letters are visible and clear in the prediction from the global model. The edges of all three letters are relatively blurry, indicating a mixture of black and white pixels on the edges. On the other hand, every letter is quite solid and there are nearly no holes inside every letter. That is, nearly all pixels inside the letters are determined as white pixels in the global model. It is clear that the FedStNN can capture the general shape of letters. In this experiment, we allow every client to learn the shape in a certain area of the image and, with FedStNN, we can successfully reproduce the image by combining all pieces of the image.

\section{Data Availability Statement}
All data in experiments \ref{exp:1D} and \ref{exp:2d} are reproducible by defining the same functions. The image in experiment \ref{exp:2d_figure} can be found at \href{https://www.google.com/imgres?q=fsu%20logo&imgurl=https%3A%2F%2Fwww.tallahassee.com%2Fgcdn%2Fauthoring%2Fauthoring-images%2F2024%2F07%2F05%2FPTAL%2F74312350007-screenshot-367.png%3Fwidth%3D1200%26disable%3Dupscale%26format%3Dpjpg%26auto%3Dwebp&imgrefurl=https%3A%2F%2Fwww.tallahassee.com%2Fstory%2Fnews%2Flocal%2Ffsu-news%2F2024%2F07%2F08%2Fflorida-state-rolls-out-new-logos-part-of-ongoing-rebranding-efforts%2F74312191007%2F&docid=KJTNl8UZOgwbFM&tbnid=tt8KXuUJKmdy7M&vet=12ahUKEwjHu7fm7dCNAxVqZzABHTkgDMMQM3oECBcQAA..i&w=1200&h=688&hcb=2&ved=2ahUKEwjHu7fm7dCNAxVqZzABHTkgDMMQM3oECBcQAA}{FSU Logo} and change its color to black and white. 












\bibliographystyle{Bibliography_Style}

\bibliography{References}

\begin{thebibliography}{38}
\expandafter\ifx\csname natexlab\endcsname\relax\def\natexlab#1{#1}\fi
\expandafter\ifx\csname url\endcsname\relax
  \def\url#1{\texttt{#1}}\fi
\expandafter\ifx\csname urlprefix\endcsname\relax\def\urlprefix{URL }\fi

\bibitem[{Archibald and Bao(2022)}]{Bao2022_KL}
Archibald, R. and Bao, F., \titlecap{Kernel learning backward SDE filter for data assimilation}, {\em J. Comput. Phys.}, vol.~{\bf 455}, no.~3, p.~111009, 2022.

\bibitem[{Archibald et~al.(2024)Archibald, Bao, Cao, and Sun}]{SNN_Convergence}
Archibald, R., Bao, F., Cao, Y., and Sun, H., \titlecap{Numerical Analysis for Convergence of a Sample-Wise Backpropagation Method for Training Stochastic Neural Networks}, {\em SIAM Journal on Numerical Analysis}, vol.~{\bf 62}, no.~2, pp.~593--621, 2024.

\bibitem[{Archibald et~al.(2020{\natexlab{a}})Archibald, Bao, Cao, and Zhang}]{archibald2021backwardsdemethoduncertainty}
Archibald, R., Bao, F., Cao, Y., and Zhang, H., \titlecap{Uncertainty Quantification in Deep Learning through Stochastic Maximum Principle}, {\em CoRR}, vol.~{\bf abs/2011.14145}, 2020{\natexlab{a}}.
\newline\urlprefix\url{https://arxiv.org/abs/2011.14145}

\bibitem[{Archibald et~al.(2020{\natexlab{b}})Archibald, Bao, and Yong}]{Bao_EAJAM20}
Archibald, R., Bao, F., and Yong, J., \titlecap{A Stochastic Gradient Descent Approach for Stochastic Optimal Control}, {\em East Asian Journal on Applied Mathematics}, vol.~{\bf 10}, no.~4, 2020{\natexlab{b}}.

\bibitem[{Archibald et~al.(2020{\natexlab{c}})Archibald, Bao, Yong, and Zhou}]{Bao_Control_20}
Archibald, R., Bao, F., Yong, J., and Zhou, T., \titlecap{An efficient numerical algorithm for solving data driven feedback control problems}, {\em Journal of Scientific Computing}, vol.~{\bf 85}, no.~51, 2020{\natexlab{c}}.

\bibitem[{Bao et~al.(2020)Bao, Cao, and Han}]{Bao_Control2020}
Bao, F., Cao, Y., and Han, X., \titlecap{Forward backward doubly stochastic differential equations and optimal Filtering of diffusion processes}, {\em Communications in Mathematical Sciences}, vol.~{\bf 18}, no.~3, pp.~635--661, 2020.

\bibitem[{Bao et~al.(2016)Bao, Cao, Meir, and Zhao}]{Bao_first}
Bao, F., Cao, Y., Meir, A., and Zhao, W., \titlecap{A first order scheme for backward doubly stochastic differential equations}, {\em SIAM/ASA J. Uncertain. Quantif.}, vol.~{\bf 4}, no.~1, pp.~413--445, 2016.
\newline\urlprefix\url{https://doi.org/10.1137/14095546X}

\bibitem[{Bao et~al.(2018)Bao, Cao, and Zhao}]{Bao2018}
Bao, F., Cao, Y., and Zhao, W., \titlecap{A backward doubly stochastic differential equation approach for nonlinear filtering problems}, {\em Commun. Comput. Phys.}, vol.~{\bf 23}, no.~5, pp.~1573--1601, 2018.

\bibitem[{Bao and Maroulas(2017)}]{BSDE_filter}
Bao, F. and Maroulas, V., \titlecap{Adaptive meshfree backward {SDE} filter}, {\em SIAM J. Sci. Comput.}, vol.~{\bf 39}, no.~6, pp.~A2664--A2683, 2017.
\newline\urlprefix\url{https://doi.org/10.1137/16M1100277}

\bibitem[{Bonawitz et~al.(2017)Bonawitz, Ivanov, Kreuter, Marcedone, McMahan, Patel, Ramage, Segal, and Seth}]{Bonawitz}
Bonawitz, K., Ivanov, V., Kreuter, B., Marcedone, A., McMahan, H.B., Patel, S., Ramage, D., Segal, A., and Seth, K., \titlecap{Practical Secure Aggregation for Privacy-Preserving Machine Learning}, {\em Proceedings of the 2017 ACM SIGSAC Conference on Computer and Communications Security}, CCS '17, Association for Computing Machinery, New York, NY, USA, p.~1175–1191, 2017.
\newline\urlprefix\url{https://doi.org/10.1145/3133956.3133982}

\bibitem[{Chellapandi et~al.(2023)Chellapandi, Yuan, Brinton, {\.Z}ak, and Wang}]{FL_Autodriving_application}
Chellapandi, V.P., Yuan, L., Brinton, C.G., {\.Z}ak, S.H., and Wang, Z., \titlecap{Federated learning for connected and automated vehicles: A survey of existing approaches and challenges}, {\em IEEE Transactions on Intelligent Vehicles}, vol.~{\bf 9}, no.~1, pp.~119--137, 2023.

\bibitem[{Chen and Chao(2020)}]{chen2021fedbemakingbayesianmodel}
Chen, H. and Chao, W., \titlecap{FedDistill: Making Bayesian Model Ensemble Applicable to Federated Learning}, {\em CoRR}, vol.~{\bf abs/2009.01974}, 2020.
\newline\urlprefix\url{https://arxiv.org/abs/2009.01974}

\bibitem[{Dean and Ghemawat(2004)}]{mapreduce}
Dean, J. and Ghemawat, S., \titlecap{MapReduce: Simplified Data Processing on Large Clusters}, {\em OSDI'04: Sixth Symposium on Operating System Design and Implementation}, San Francisco, CA, pp.~137--150, 2004.

\bibitem[{Dwork et~al.(2006)Dwork, McSherry, Nissim, and Smith}]{differential_privacy}
Dwork, C., McSherry, F., Nissim, K., and Smith, A., \titlecap{Calibrating noise to sensitivity in private data analysis}, {\em Proceedings of the Third Conference on Theory of Cryptography}, TCC'06, Springer-Verlag, Berlin, Heidelberg, p.~265–284, 2006.
\newline\urlprefix\url{https://doi.org/10.1007/11681878\_14}

\bibitem[{Fachola et~al.(2023)Fachola, Tornaría, Bermolen, Capdehourat, Etcheverry, and Fariello}]{FL_Education_application}
Fachola, C., Tornaría, A., Bermolen, P., Capdehourat, G., Etcheverry, L., and Fariello, M.I., \titlecap{Federated Learning for Data Analytics in Education}, {\em Data}, vol.~{\bf 8}, no.~2, 2023.
\newline\urlprefix\url{https://www.mdpi.com/2306-5729/8/2/43}

\bibitem[{Gentry(2009)}]{homomorphic_encryption}
Gentry, C., \titlecap{Fully homomorphic encryption using ideal lattices}, {\em Proceedings of the Forty-First Annual ACM Symposium on Theory of Computing}, STOC '09, Association for Computing Machinery, New York, NY, USA, p.~169–178, 2009.
\newline\urlprefix\url{https://doi.org/10.1145/1536414.1536440}

\bibitem[{Hardy et~al.(2017)Hardy, Henecka, Ivey{-}Law, Nock, Patrini, Smith, and Thorne}]{hardy2017privatefederatedlearningvertically}
Hardy, S., Henecka, W., Ivey{-}Law, H., Nock, R., Patrini, G., Smith, G., and Thorne, B., \titlecap{Private federated learning on vertically partitioned data via entity resolution and additively homomorphic encryption}, {\em CoRR}, vol.~{\bf abs/1711.10677}, 2017.
\newline\urlprefix\url{http://arxiv.org/abs/1711.10677}

\bibitem[{He et~al.(2021)He, Ceyani, Balasubramanian, Annavaram, and Avestimehr}]{he2021spreadgnnserverlessmultitaskfederated}
He, C., Ceyani, E., Balasubramanian, K., Annavaram, M., and Avestimehr, S., \titlecap{SpreadGNN: Serverless Multi-task Federated Learning for Graph Neural Networks}, {\em CoRR}, vol.~{\bf abs/2106.02743}, 2021.
\newline\urlprefix\url{https://arxiv.org/abs/2106.02743}

\bibitem[{He et~al.(2019)He, Tan, Tang, Qiu, and Liu}]{he2020centralserverfreefederated}
He, C., Tan, C., Tang, H., Qiu, S., and Liu, J., \titlecap{Central Server Free Federated Learning over Single-sided Trust Social Networks}, {\em CoRR}, vol.~{\bf abs/1910.04956}, 2019.
\newline\urlprefix\url{http://arxiv.org/abs/1910.04956}

\bibitem[{Joshi et~al.(2022)Joshi, Pal, and Sankarasubbu}]{FL_Health_application}
Joshi, M., Pal, A., and Sankarasubbu, M., \titlecap{Federated Learning for Healthcare Domain - Pipeline, Applications and Challenges}, {\em ACM Transactions on Computing for Healthcare}, vol.~{\bf 3}, no.~4, p.~1–36, 2022.
\newline\urlprefix\url{http://dx.doi.org/10.1145/3533708}

\bibitem[{Jospin et~al.(2022)Jospin, Laga, Boussaid, Buntine, and Bennamoun}]{BNN_intro}
Jospin, L.V., Laga, H., Boussaid, F., Buntine, W., and Bennamoun, M., \titlecap{Hands-On Bayesian Neural Networks—A Tutorial for Deep Learning Users}, {\em IEEE Computational Intelligence Magazine}, vol.~{\bf 17}, no.~2, p.~29–48, 2022.
\newline\urlprefix\url{http://dx.doi.org/10.1109/MCI.2022.3155327}

\bibitem[{Kerkouche et~al.(2021)Kerkouche, \'{A}cs, Castelluccia, and Genev\`{e}s}]{FL_realtimehealth_application}
Kerkouche, R., \'{A}cs, G., Castelluccia, C., and Genev\`{e}s, P., \titlecap{Privacy-preserving and bandwidth-efficient federated learning: an application to in-hospital mortality prediction}, {\em Proceedings of the Conference on Health, Inference, and Learning}, CHIL '21, Association for Computing Machinery, New York, NY, USA, p.~25–35, 2021.
\newline\urlprefix\url{https://doi.org/10.1145/3450439.3451859}

\bibitem[{Khan et~al.(2024)Khan, S{\'a}nchez, and Domingo-Ferrer}]{FL_NLP_application}
Khan, Y., S{\'a}nchez, D., and Domingo-Ferrer, J., \titlecap{Federated learning-based natural language processing: a systematic literature review}, {\em Artificial Intelligence Review}, vol.~{\bf 57}, no.~12, p.~320, 2024.
\newline\urlprefix\url{https://doi.org/10.1007/s10462-024-10970-5}

\bibitem[{Koetsier et~al.(2022)Koetsier, Fiosina, Gremmel, Müller, Woisetschläger, and Sester}]{FL_Trajectory_application}
Koetsier, C., Fiosina, J., Gremmel, J.N., Müller, J.P., Woisetschläger, D.M., and Sester, M., \titlecap{Detection of anomalous vehicle trajectories using federated learning}, {\em ISPRS Open Journal of Photogrammetry and Remote Sensing}, vol.~{\bf 4}, p.~100013, 2022.
\newline\urlprefix\url{https://www.sciencedirect.com/science/article/pii/S2667393222000023}

\bibitem[{Lalitha et~al.(2019)Lalitha, Kilinc, Javidi, and Koushanfar}]{lalitha2019peertopeerfederatedlearninggraphs}
Lalitha, A., Kilinc, O.C., Javidi, T., and Koushanfar, F., \titlecap{Peer-to-peer Federated Learning on Graphs}, {\em CoRR}, vol.~{\bf abs/1901.11173}, 2019.
\newline\urlprefix\url{http://arxiv.org/abs/1901.11173}

\bibitem[{Li et~al.(2019)Li, Huang, Yang, Wang, and Zhang}]{li2020convergencefedavgnoniiddata}
Li, X., Huang, K., Yang, W., Wang, S., and Zhang, Z., \titlecap{On the convergence of fedavg on non-iid data}, {\em arXiv preprint arXiv:1907.02189}, 2019.

\bibitem[{Liang et~al.(2024)Liang, Sun, Archibald, and Bao}]{Liang_Control}
Liang, S., Sun, H., Archibald, R., and Bao, F., \titlecap{Convergence Analysis for an Online Data-Driven Feedback Control Algorithm}, {\em Mathematics}, vol.~{\bf 12}, 2024.

\bibitem[{Liu et~al.(2023)Liu, Tian, Chakraborty, Feng, Pei, Zhen, and Yu}]{FL_Traffic_application}
Liu, L., Tian, Y., Chakraborty, C., Feng, J., Pei, Q., Zhen, L., and Yu, K., \titlecap{Multilevel Federated Learning-Based Intelligent Traffic Flow Forecasting for Transportation Network Management}, {\em IEEE Transactions on Network and Service Management}, vol.~{\bf 20}, no.~2, pp.~1446--1458, 2023.

\bibitem[{McMahan et~al.(2016)McMahan, Moore, Ramage, and y~Arcas}]{DBLP:journals/corr/McMahanMRA16}
McMahan, H.B., Moore, E., Ramage, D., and y~Arcas, B.A., \titlecap{Federated Learning of Deep Networks using Model Averaging}, {\em CoRR}, vol.~{\bf abs/1602.05629}, 2016.
\newline\urlprefix\url{http://arxiv.org/abs/1602.05629}

\bibitem[{Nguyen et~al.(2021)Nguyen, Ding, Pathirana, Seneviratne, Li, and Vincent~Poor}]{FL_IoT_application}
Nguyen, D.C., Ding, M., Pathirana, P.N., Seneviratne, A., Li, J., and Vincent~Poor, H., \titlecap{Federated Learning for Internet of Things: A Comprehensive Survey}, {\em IEEE Communications Surveys \& Tutorials}, vol.~{\bf 23}, no.~3, p.~1622–1658, 2021.
\newline\urlprefix\url{http://dx.doi.org/10.1109/COMST.2021.3075439}

\bibitem[{Pandya et~al.(2023)Pandya, Srivastava, Jhaveri, Babu, Bhattacharya, Maddikunta, Mastorakis, Piran, and Gadekallu}]{FL_SmartCity_application}
Pandya, S., Srivastava, G., Jhaveri, R., Babu, M.R., Bhattacharya, S., Maddikunta, P.K.R., Mastorakis, S., Piran, M.J., and Gadekallu, T.R., \titlecap{Federated learning for smart cities: A comprehensive survey}, {\em Sustainable Energy Technologies and Assessments}, vol.~{\bf 55}, p.~102987, 2023.
\newline\urlprefix\url{https://www.sciencedirect.com/science/article/pii/S2213138822010359}

\bibitem[{Peng(1990)}]{peng1990general}
Peng, S., \titlecap{A general stochastic maximum principle for optimal control problems}, {\em SIAM Journal on control and optimization}, vol.~{\bf 28}, no.~4, pp.~966--979, 1990.

\bibitem[{Shokri and Shmatikov(2015)}]{DSSGD}
Shokri, R. and Shmatikov, V., \titlecap{Privacy-preserving deep learning}, {\em 2015 53rd Annual Allerton Conference on Communication, Control, and Computing (Allerton)}, pp.~909--910, 2015.

\bibitem[{Su et~al.(2022)Su, Wang, Luan, Zhang, Li, Chen, and Cao}]{FL_SmartGrid_application}
Su, Z., Wang, Y., Luan, T.H., Zhang, N., Li, F., Chen, T., and Cao, H., \titlecap{Secure and Efficient Federated Learning for Smart Grid With Edge-Cloud Collaboration}, {\em IEEE Transactions on Industrial Informatics}, vol.~{\bf 18}, no.~2, pp.~1333--1344, 2022.

\bibitem[{Sun and Bao(2021)}]{Sun_Control}
Sun, H. and Bao, F., \titlecap{Meshfree Approximation for Stochastic Optimal Control Problems}, {\em Communications in Mathematical Research}, vol.~{\bf 37}, p.~387–420, 2021.

\bibitem[{Wen et~al.(2023)Wen, Zhang, Lan, Cui, Cai, and Zhang}]{FL_Finance_application}
Wen, J., Zhang, Z., Lan, Y., Cui, Z., Cai, J., and Zhang, W., \titlecap{A survey on federated learning: challenges and applications}, {\em International Journal of Machine Learning and Cybernetics}, vol.~{\bf 14}, no.~2, pp.~513--535, 2023.
\newline\urlprefix\url{https://doi.org/10.1007/s13042-022-01647-y}

\bibitem[{Wu et~al.(2021)Wu, Jiang, Han, Yuan, Li, and Zhang}]{FL_Motion_applicaiton}
Wu, T., Jiang, M., Han, Y., Yuan, Z., Li, X., and Zhang, L., \titlecap{A Traffic-Aware Federated Imitation Learning Framework for Motion Control at Unsignalized Intersections with Internet of Vehicles}, {\em Electronics}, vol.~{\bf 10}, no.~24, 2021.
\newline\urlprefix\url{https://www.mdpi.com/2079-9292/10/24/3050}

\bibitem[{Xie et~al.(2023)Xie, Li, Zhou, and Dong}]{FL_License_application}
Xie, R., Li, C., Zhou, X., and Dong, Z., \titlecap{Asynchronous Federated Learning for Real-Time Multiple Licence Plate Recognition Through Semantic Communication}, {\em ICASSP 2023 - 2023 IEEE International Conference on Acoustics, Speech and Signal Processing (ICASSP)}, pp.~1--5, 2023.

\end{thebibliography}
\end{document}